\documentclass{article} % For LaTeX2e
\usepackage{iclr2026_conference,times}
\usepackage{algorithm}
\usepackage{algorithmic}
\usepackage{amsmath} 
\usepackage{colortbl}
\usepackage{multirow}
\usepackage{diagbox}
\usepackage{xcolor}
\usepackage{subcaption}
\usepackage{graphicx}
\usepackage{booktabs}
\newcommand{\revise}[1]{#1}

\usepackage{amsmath,amsfonts,bm}

\def\eqref#1{equation~\ref{#1}}
\def\1{\bm{1}}

\DeclareMathAlphabet{\mathsfit}{\encodingdefault}{\sfdefault}{m}{sl}
\SetMathAlphabet{\mathsfit}{bold}{\encodingdefault}{\sfdefault}{bx}{n}

\usepackage[colorlinks=true, linkcolor=blue, urlcolor=blue, citecolor=blue]{hyperref}
\usepackage{url}

\title{SSD-GS: Scattering and Shadow Decomposition for Relightable 3D Gaussian Splatting}

\author{Iris Zheng, Guojun Tang, Alexander Doronin, Paul Teal, Fang-Lue Zhang\thanks {Fang-Lue Zhang (fanglue.zhang@vuw.ac.nz) is the corresponding author} \\
Victoria University of Wellington }

\iclrfinalcopy % Uncomment for camera-ready version, but NOT for submission.
\begin{document}

\maketitle

\begin{figure*}[h]
    \centering
    \includegraphics[width=\linewidth]{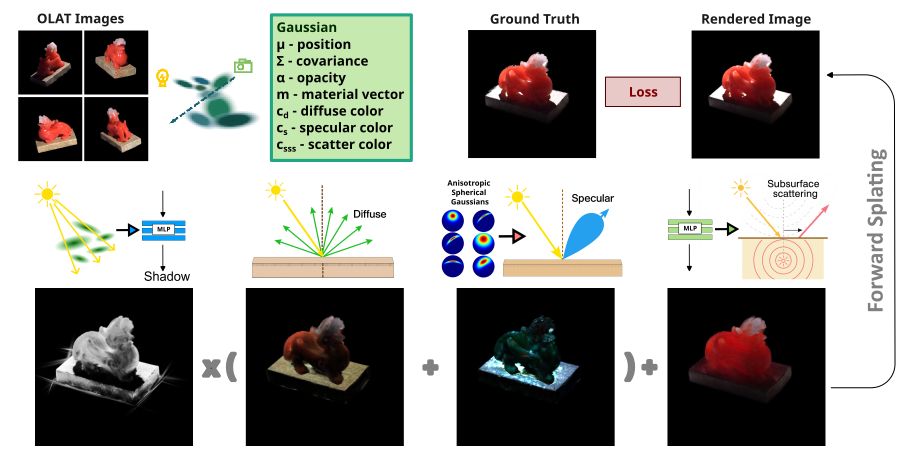}
    \caption{Overview of the proposed SSD-GS pipeline. Our method incorporates four physically inspired reflectance terms: diffuse, specular, shadow, and subsurface scattering, to model realistic light–material interactions. These components are progressively introduced during training, allowing the network to gradually disentangle complex illumination effects and improve relighting fidelity under unseen lighting conditions.}
    \label{fig:framework}
\end{figure*}

\begin{abstract}
We present SSD-GS, a physically-based relighting framework built upon 3D Gaussian Splatting (3DGS) that achieves high-quality reconstruction and photorealistic relighting under novel lighting conditions. In physically-based relighting, accurately modeling light-material interactions is essential for faithful appearance reproduction. However, existing 3DGS-based relighting methods adopt coarse shading decompositions, either modeling only diffuse and specular reflections or relying on neural networks to approximate shadows and scattering. This leads to limited fidelity and poor physical interpretability, particularly for anisotropic metals and translucent materials. To address these limitations, SSD-GS decomposes reflectance into four components: diffuse, specular, shadow, and subsurface scattering. We introduce a learnable dipole-based scattering module for subsurface transport, an occlusion-aware shadow formulation that integrates visibility estimates with a refinement network, and an enhanced specular component with an anisotropic Fresnel-based model. Through progressive integration of all components during training, SSD-GS effectively disentangles lighting and material properties, even for unseen illumination conditions, as demonstrated on the challenging OLAT dataset. Experiments demonstrate superior quantitative and perceptual relighting quality compared to prior methods and pave the way for downstream tasks, including controllable light source editing and interactive scene relighting. The source code is available at: \href{https://github.com/irisfreesiri/SSD-GS}{https://github.com/irisfreesiri/SSD-GS}.
\end{abstract}

\section{Introduction}

Photorealistic 3D reconstruction with relightable capabilities has become increasingly important across domains such as AR/VR for digital humans, cinematic visual effects, cultural heritage preservation, and medical simulation. Traditional methods \citep{levoy_light_1996, seitz_view_1996, seitz_photorealistic_1997, snavely_photo_2006}, however, typically compromise either geometric precision or photorealistic quality, particularly in complex lighting conditions or with reflective and textured surfaces. While these approaches enabled view synthesis under captured illumination, they relied on explicit geometric reconstructions and provided no means to disentangle reflectance from lighting. As a result, they cannot support relighting under novel illumination, which is essential for realistic appearance reproduction in many applications. More recently, neural rendering approaches, in particular those based on Neural Radiance Fields (NeRF) \citep{mildenhall_nerf_2020}, have made notable progress by jointly encoding geometry and appearance in an implicit volumetric representation. Methods such as DNL \citep{gao_deferred_2020} and NRHints \citep{zeng_relighting_2023} introduce explicit lighting supervision and learnable shading representations to support relightable view synthesis. However, NeRF-based methods typically suffer from high computational cost, which limits their practicality for interactive or real-time applications. 

3D Gaussian Splatting (3DGS), initially developed for real-time radiance field rendering, has emerged as a compelling alternative to NeRF-style implicit representations that rely on ray marching, offering superior computational efficiency and rendering fidelity. Recent extensions of 3DGS for relightable rendering fall into two main paradigms. Some methods assume static lighting conditions during training~\citep{jiang_gaussianshader_2023, liang_gs-ir_2024, chen_gi-gs_2024, gao_relightable_2024}, which fundamentally lacks their flexibility for photorealistic relighting. Others leverage dynamic lighting configurations such as one-light-at-a-time (OLAT) capture setups~\citep{bi_gs3_2024, kuang_olat_2024, fan_rng_2025, dihlmann_subsurface_2025}, offering more physically plausible supervision but making it difficult to disentangle material properties from illumination. This disentanglement is crucial for simulating complex light transport behaviors of real-world materials, where nonlinear interactions give rise to visually critical phenomena such as gradient soft shadows and subsurface scattering. Consequently, developing robust techniques to model these intricate lighting-material interactions remains a substantial technical challenge for relightable 3D reconstruction.

We propose \textbf{SSD-GS}, a physically-based relighting method designed for 3DGS\revise{, where “physically-based” follows the real-time PBR convention and denotes the use of physically inspired reflectance models within an efficient rasterized framework}. Built upon the 3DGS pipeline, our framework explicitly decomposes complex reflectance into four components: diffuse, specular, subsurface scattering, and shadow. Our main contributions are:

\begin{itemize}  
    \item We introduce a learnable dipole-based scattering module that simulates realistic subsurface \revise{scattering effects} using \revise{physically motivated} diffusion profiles.
  
    \item We design an occlusion-aware shadow formulation that combines a visibility prior with a learned refinement network, enabling accurate modeling of view- and light-dependent shadowing effects.
  
    \item We progressively integrate all reflectance components (diffuse, specular, shadow, and subsurface scattering) during training and refine both lighting and camera conditions, leading to improved relighting quality and stronger generalization under novel illuminations.
\end{itemize}

\section{Related Works}

Accurate relighting and novel view synthesis require recovering both scene geometry and material appearance under illumination. We review NeRF- and 3DGS-based relighting methods, followed by subsurface scattering (SSS) models for physically plausible rendering.

\textbf{NeRF-based Relighting.}
Neural Radiance Fields (NeRF)\citep{mildenhall_nerf_2020} represent scenes as volumetric fields optimized from posed RGB images, enabling photorealistic novel view synthesis under fixed lighting. Extensions for relighting factorize appearance into reflectance and illumination using priors or explicit transport modeling. For instance, NeRV\citep{srinivasan_nerv_2021}, NeRD~\citep{boss_nerd_2021}, and NeRFactor~\citep{zhang_nerfactor_2021} disentangle reflectance under static lighting with geometry-aware priors, while PhySG~\citep{zhang_physg_2021} uses spherical Gaussians to represent BRDFs and environment lighting.
To address directional lighting, ReNeRF~\citep{xu_renerf_2023} models near-field OLAT illumination via a spherical codebook and light transport decoder, enabling spatially varying lighting. NRHints\citep{zeng_relighting_2023} injects OLAT-derived shadow and highlight hints into a NeRF-style radiance field, achieving relighting effects comparable to DNL\citep{gao_deferred_2020} but using a fully volumetric, single-branch design.
However, NeRF-based methods suffer from implicit, non-physical representations, making decomposition hard to interpret or control. Moreover, they are computationally expensive, requiring hours of training per scene.

\textbf{Gaussian-based Relighting.}
3DGS~\citep{kerbl_3d_2023} models scenes as point clouds of anisotropic Gaussians with learned extent, opacity, and view-dependent appearance. While efficient for view synthesis, its SH-based color encoding~\citep{ramamoorthi_efficient_2001, sloan_precomputed_2002} is inherently limited to smooth, low-frequency angular variations, which reduces expressiveness for capturing high-frequency effects such as specular highlights and scattering.
Several extensions enhance 3DGS with physically motivated components, including GaussianShader~\citep{jiang_gaussianshader_2023}, GI-GS~\citep{chen_gi-gs_2024}, and R3DG~\citep{gao_relightable_2024}. However, these typically assume static lighting conditions, which prevents them from generalizing to novel illuminations. Their relightable variants usually perform global relighting using environmental maps, but lack the ability to model precise changes in individual light sources.
To overcome these limitations, recent works exploit dynamic lighting conditions, most notably one-light-at-a-time (OLAT) datasets. GS$^3$~\citep{bi_gs3_2024} decomposes reflectance by modeling diffuse and specular terms at the Gaussian level, while handling shadows and other residual effects at the pixel level in a deferred rendering style~\citep{ye_3d_2024}. However, this design struggles to capture complex light transport phenomena such as soft shadows and indirect illumination. OLAT Gaussians~\citep{kuang_olat_2024} use directional encodings with two MLPs to model incident and scattering components, but their use of a proxy mesh for normal supervision makes them highly sensitive to the quality of the proxy geometry. RNG~\citep{fan_rng_2025} achieves improved shadow quality by introducing a latent appearance code, which replaces physically meaningful shading representations and thus sacrifices interpretability. Inspired by these OLAT-based approaches, we introduce physically interpretable shading to better disentangle lighting–material interactions and extend performance to more diverse datasets.

\begin{figure}[h]
  \centering
  \includegraphics[width=0.8\linewidth]{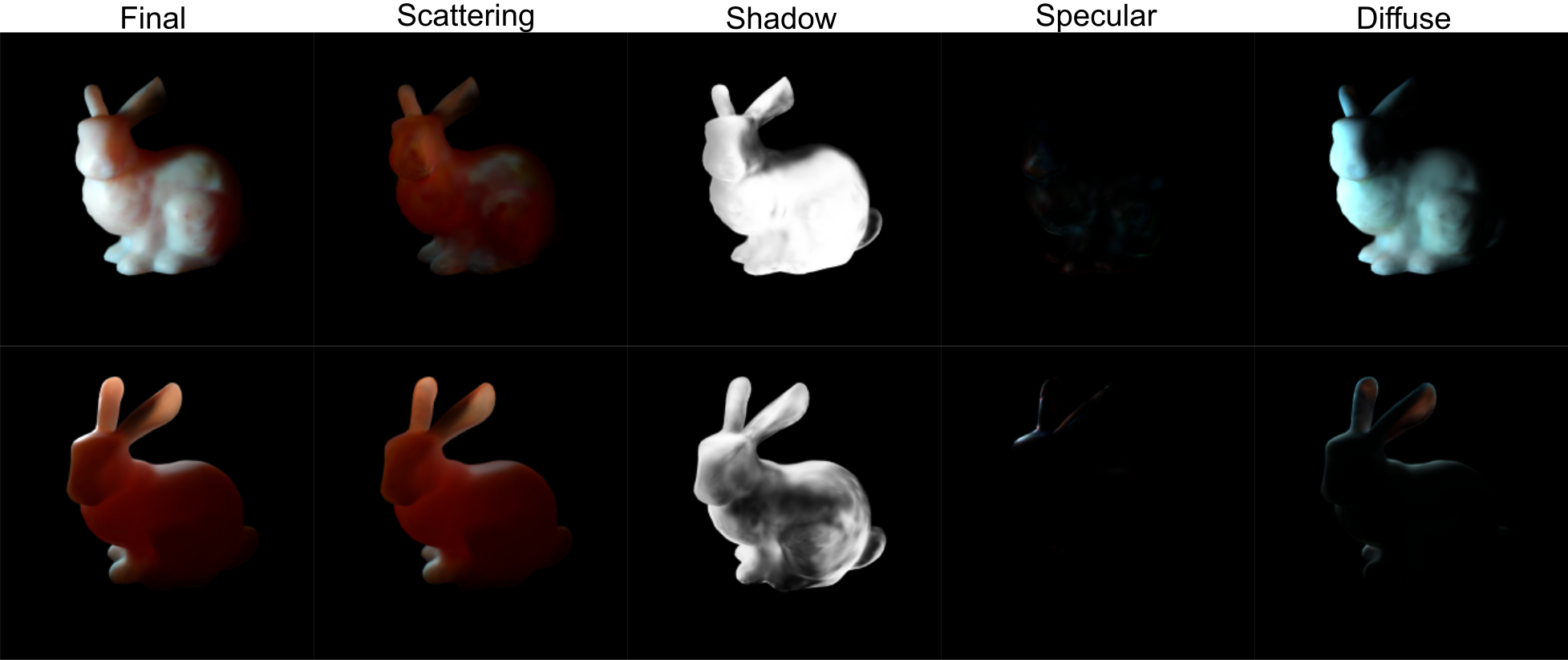}
  \caption{Relighting results from our SSD-GS pipeline. The same \textit{Bunny} view under two different lighting conditions from the SSS-GS synthetic dataset~\citep{dihlmann_subsurface_2025}.}
  \label{fig:teaser}
\end{figure}

\textbf{Subsurface Scattering. }
Subsurface scattering (SSS) has been extensively studied for simulating light transport in translucent materials such as skin, jade, wax, and marble. Classical approaches, including the standard dipole~\citep{jensen_practical_2001}, quantized dipole~\citep{deon_quantized-diffusion_2011}, and directional dipole~\citep{frisvad_directional_2014}, offer efficient and realistic approximations. Extensions such as shape-adaptive dipole models~\citep{vicini_learned_2019} 
and advanced BSSRDF formulations~\citep{yan_bssrdf_2017} 
further enhance accuracy and generality. More recently, subsurface scattering has been explored in neural rendering frameworks through learning-based techniques. Neural SSS~\citep{tg_neural_2024} approximates the translucent appearance using per-view and per-light neural reflectance fields, but it relies heavily on dense supervision and lacks physical interpretability. In the context of Gaussian Splatting, SSS-GS~\citep{dihlmann_subsurface_2025} directly learns the subsurface scattering radiance via a neural network conditioned on Gaussian and lighting inputs. The output is blended with BRDF shading using a learned weight, treating SSS as a residual term \revise{rather than a physically motivated subsurface model}.
In contrast, we integrate a physically grounded subsurface scattering approach into the 3DGS pipeline, based on the standard dipole diffusion approximation~\citep{jensen_practical_2001}. This classical method provides a closed-form \revise{BSSRDF that approximates multiple scattering} in homogeneous media. By embedding it into the Gaussian Splatting framework, we enable efficient, interpretable simulation of soft scattering effects, while maintaining modular compatibility with other shading components such as diffuse, specular, and shadow terms.

\section{Preliminary}
Our method builds on the 3D Gaussian Splatting (3DGS) framework~\citep{kerbl_3d_2023}, which represents a scene as a set of anisotropic 3D Gaussians. Each Gaussian is defined by its center \( \mathbf{x}_i \), opacity \( \alpha_i \), and a covariance matrix \( \Sigma_i \). The covariance matrix is parameterized via a rotation matrix \( R_i \) and a scaling matrix \( S_i \), such that \( \Sigma_i = R_i S_i S_i^\top R_i^\top \). During rendering, the Gaussians are projected onto the image plane and composited using front-to-back alpha blending as:

\begin{equation}
\mathbf{C}_{\text{pixel}} = \sum_{i=1}^{N} T_i \cdot \alpha_i \cdot \mathbf{C}_i, \quad T_i = \prod_{j=1}^{i-1} (1 - \alpha_j)
\end{equation}

Each Gaussian color \( \mathbf{C}_i \) is computed using a view-dependent SH expansion:%~\citep{ramamoorthi_efficient_2001}:
\begin{equation}
\mathbf{C}_i(\mathbf{v}) = \sum_{b=1}^{B} \mathbf{c}_{i,b} \cdot Y_b(\mathbf{v})
\end{equation}

While effective for encoding smooth appearance, this SH-based model lacks physical grounding and struggles to capture high-frequency view-dependent effects. In this work, we replace it with a decomposed physically-based  model to better capture full-frequency light–material interactions.

\section{Methodology}
We extend the 3D Gaussian Splatting (3DGS) framework by incorporating a physically-based reflectance model that replaces its original spherical harmonics (SH)-based appearance representation. Our formulation decomposes shading into four components—diffuse, specular, shadow, and subsurface scattering (SSS)—each modeled either analytically or using lightweight neural fields. These components are evaluated per-Gaussian and composited to form the final image, enabling interpretable supervision and relightable rendering under novel illumination. Their visual effects are illustrated in Fig.~\ref{fig:teaser}, and a detailed ablation study is provided in Sec.~\ref{section:ablation-study-components}. An overview of the formulation is illustrated in Fig.~\ref{fig:framework}.

\subsection{PBR-Based Shading}
We formulate a physically-based shading function that operates directly on the 3D Gaussian representation. Unlike prior work that employs view-dependent spherical harmonics (SH) for color synthesis~\citep{kerbl_3d_2023}, we decompose reflectance into physically interpretable terms, enabling improved photorealism, per-component supervision, and controllable relighting. The color of each Gaussian is computed as:
\begin{equation}
\mathbf{C}_i = \left(c_d f_d + c_s f_s \right) \cdot S(\mathbf{x}) + c_{sss} f_{sss}
\label{eq:shading_equation}
\end{equation}
where: \( f_d, f_s, f_{sss} \) denote the scalar reflectance intensities for diffuse, specular, and subsurface scattering, defined in Eqs.~\ref{eq:diffuse}, \ref{eq:specular}, and \ref{eq:scattering}, respectively; \( c_d, c_s, c_{sss} \in \mathbb{R}^3 \) are the corresponding learned base colors for each reflectance term; \( S(\mathbf{x}) \) denotes the soft shadow factor, computed as a density-weighted average over shadow rays and further refined using an MLP, with its detailed formulation given in Eq.~\ref{eq:shadow}.
This decomposition is evaluated per Gaussian and composited through the 3DGS forward-rendering pipeline, where alpha blending accumulates Gaussian contributions into the final pixel color. The resulting image is supervised with a pixel-wise loss against the ground truth. %An overview of the complete pipeline is shown in Fig.~\ref{fig:framework}.%
\revise{A detailed analysis of the interaction between the shadow term and subsurface scattering is provided in Appendix~\ref{appendix:shadow-sss-interaction}.}

\subsection{Subsurface Scattering Term}
We model subsurface scattering (SSS) using the standard dipole diffusion profile, with scattering properties defined per Gaussian. To predict these parameters, we train a neural field \(\Theta_{\text{SSS}}\) that maps spatial and directional inputs to the corresponding scattering coefficients.

\begin{equation}
\{\sigma_s, \sigma_a, r\} = \Theta_{\text{SSS}}(\mathbf{x} \mid \mathbf{w}_o, \mathbf{w}_i, \mathbf{n}, \mathbf{m})
\end{equation}
where \( \mathbf{x} \) denotes the Gaussian center, \( \omega_i \) and \( \omega_o \) are the light and view directions, \( \mathbf{n} \) is the surface normal derived from its local z-axis, and \( \mathbf{m} \in \mathbb{R}^6 \) is a learnable per-Gaussian material embedding; $\sigma_s$, $\sigma_a$, and $r$ denote the scattering coefficient, absorption coefficient, and surface separation distance used in the dipole formulation.

The subsurface scattering (SSS) predictor is implemented as a 6-layer MLP with a hidden size of 256 and ReLU activations. It takes as input the positional encodings (with \( L = 4 \) frequency bands) of the spatial location \( \mathbf{x} \), the viewing direction \( \omega_o \), and the lighting direction \( \omega_i \), as well as the local surface normal \( \mathbf{n} \) and a per-Gaussian material embedding \( \mathbf{m} \). To ensure physical plausibility and improve training stability, the network outputs are passed through sigmoid activations and rescaled to fall within plausible material-specific ranges: \( \sigma_s, \sigma_a \in [0.05, 2.05] \), and \( r \in [0.1, 3.1] \).
We evaluate the standard dipole diffusion profile~\citep{jensen_practical_2001} as:

\begin{equation}
f_{sss}(r) = \frac{\alpha'}{4\pi} \left( z_r(\sigma_t d_r + 1)\frac{e^{-\sigma_t d_r}}{d_r^3} + z_r z_v (\sigma_t d_r + 1)\frac{e^{-\sigma_t d_v}}{d_v^3} \right)
\label{eq:scattering}
\end{equation}
where \( \alpha' = \frac{\sigma_s}{\sigma_s + \sigma_a} \), \( \sigma_t = \sigma_s + \sigma_a \), $z_r$ and $z_v$ are the depths of the real and virtual dipole sources, determined by the optical parameters $(\sigma_s,\sigma_a,\eta)$, and \( d_r, d_v \) are the corresponding distances from the shading point to the real and virtual dipole sources, computed from the surface separation $r$.

Our SSS formulation combines physically grounded modeling with learnable parameter prediction, enabling realistic reproduction of subsurface scattering effects without requiring external geometry~\citep{kuang_olat_2024}. Because surface normals and material properties are inferred directly from the Gaussian representation, the system remains robust under challenging geometric conditions. As a result, it generalizes well to complex or noisy regions where mesh-derived normals may be unreliable, thus preserving effective scattering estimation.

\subsection{Shadow Term}
We model soft shadows using a two-stage approach that combines \revise{per-ray shadow evaluation} with neural refinement. \revise{In the first stage, for each Gaussian we trace a shadow ray from the light source to every pixel and accumulate transmittance into visibility cues. In the second stage, a compact neural module takes these cues, together with geometry and material features, and predicts a scalar decay factor used in shading.}

\paragraph{Stage 1: Shadow Evaluation.}
\revise{Given a light direction~$\omega_i$, each Gaussian considers the set of pixels~$i$ covered by its 2D projection. For each pixel, we evaluate a shadow ray from the light source toward that pixel and accumulate the opacity of intervening Gaussians. This yields a continuous per-ray transmittance}
\revise{\begin{equation}
v_i = \prod_{k \in \mathcal{O}_i} (1 - \alpha_k),
\label{eq:shadow_transmittance}
\end{equation}}
\revise{where $\mathcal{O}_i$ is the depth-ordered set of Gaussians intersected by the shadow ray, and $\alpha_k \in [0,1]$ denotes the opacity of Gaussian~$k$.}

\revise{To obtain a soft shadow estimate, these per-ray transmittance values are aggregated using the Gaussian’s projected density as weights. Let $\rho_i$ denote the projected density of Gaussian~$g$ at pixel~$i$. The coarse visibility of Gaussian~$g$ is then defined as the density-weighted expectation, }
\begin{equation}
\hat v_g =
\frac{\sum_i \rho_i\, v_i}{\sum_i \rho_i},
\label{eq:shadow_visibility}
\end{equation}
\revise{which summarizes how much light from the direction~$\omega_i$ reaches the
Gaussian~$g$ after accounting for overlapping geometry, and serves as a compact visibility estimate.}

\paragraph{Stage 2: Neural Refinement.}
\revise{The coarse visibility ~$\hat v$ captures the primary directional shadowing trend but may miss fine variations arising from contact shadows, geometric details, and material-dependent attenuation. To account for these effects, we refine~$\hat v$ using a lightweight neural module~$\Theta_{\text{shad}}$, which predicts a shadow attenuation term as a function of position:}

\begin{equation}
S(\mathbf{x}) = \Theta_{\text{shad}}(\mathbf{x} \mid \hat{v}, \omega_i, \mathbf{m})
\label{eq:shadow}
\end{equation}

The shadow refinement network is implemented using a 3-layer MLP with 32 hidden units per layer and ReLU activations. Its inputs include the Gaussian center \( \mathbf{x} \), incident light direction \( \omega_i \), coarse shadow estimate \( \hat{v} \), and material embedding \( \mathbf{m} \). To capture high-frequency spatial and directional variation, both \( \mathbf{x} \) and \( \omega_i \) are encoded using positional encoding with \( L = 3 \) frequency levels.

\noindent
\revise{The refined shadow term~$S(\mathbf{x})$ modulates the diffuse
and specular components of our shading model, while the scattering term
is added separately. This produces the final illumination contribution
for Gaussian~$g$ as defined in Eq.~\ref{eq:shading_equation}.}

\begin{figure}[h]
    \centering
    \includegraphics[width=\linewidth]{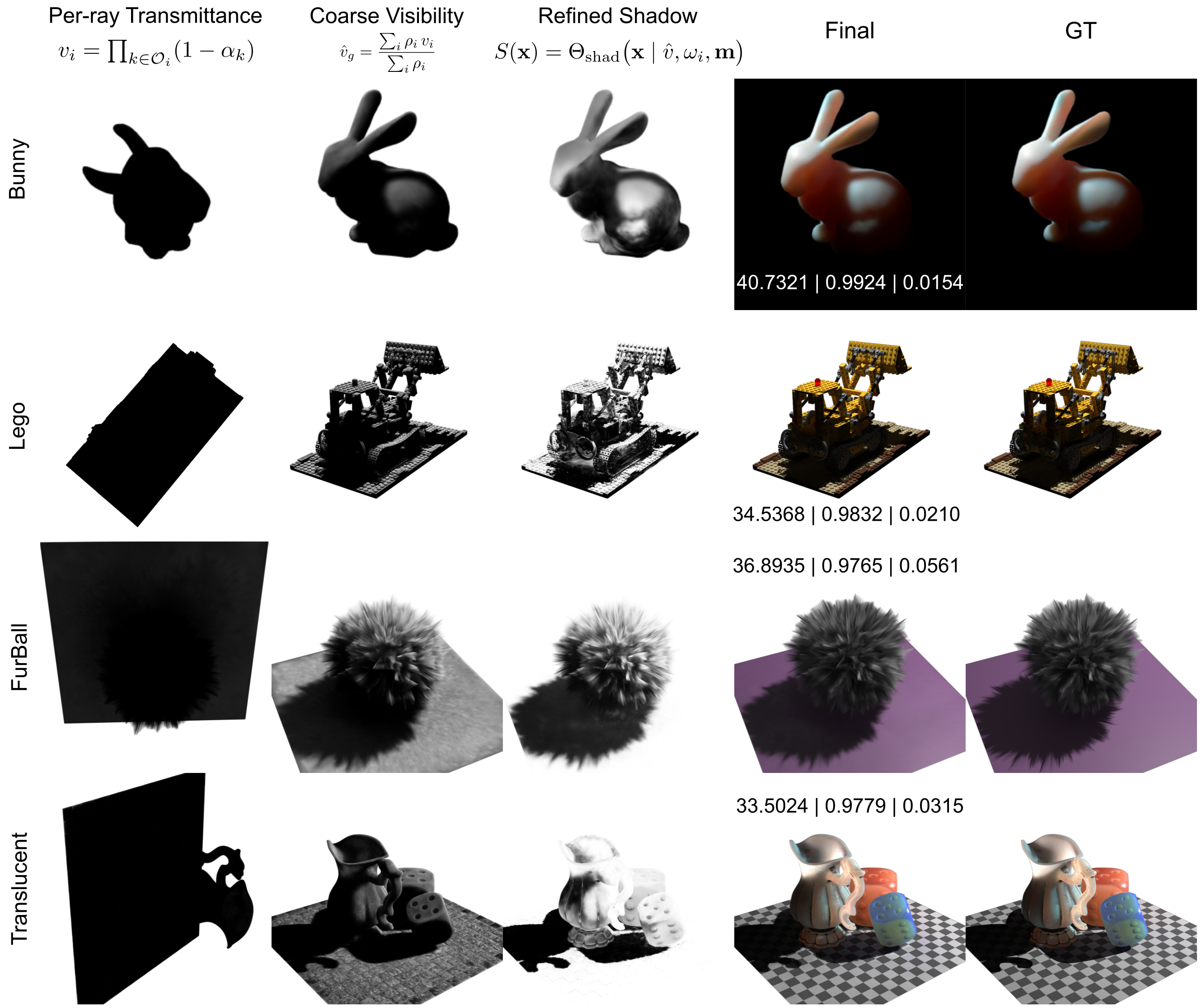}
    \caption{Shadow pipeline visualization. For each scene, we show per-ray transmittance $v_i$, coarse visibility $\hat v_g$, the refined shadow $S(x)$, and the final rendered result with metrics. The progression illustrates how continuous transmittance yields smooth, geometry-aware soft shadows.}
    \label{fig:shadow_analysis_pipeline}
\end{figure}

\revise{A visual illustration of this progression from per-ray transmittance $v_i$, to coarse visibility $\hat v_g$, and finally to the refined shadow $S(\mathbf{x})$ is provided in Fig.~\ref{fig:shadow_analysis_pipeline}. The figure highlights how continuous volumetric visibility naturally produces smooth, geometry-consistent soft shadows under point-light illumination. 
A complementary comparison against screen-space shadow accumulation methods is included in Appendix~\ref{appendix:comparison-screen-space-shadow-baselines}.}

\subsection{Diffuse and Specular Terms}
We decompose direct shading into diffuse and specular components. The diffuse term is modeled with a Lambertian BRDF, which assumes uniform surface reflectance and produces view-independent, cosine-weighted reflection. Although simple, this model provides a stable foundation for capturing low-frequency appearance and ensures physically meaningful supervision during the early stages of training. 
The specular term, in contrast, accounts for high-frequency, view-dependent reflections. We represent it as a Fresnel factor~\citep{schlick_inexpensive_1994} modulated by anisotropic spherical Gaussian (ASG) bases~\citep{xu_anisotropic_2013}. The Fresnel term captures the angular dependence of reflection intensity, particularly the sharp increase near grazing angles, while the ASG bases provide a compact yet expressive representation of anisotropic highlights. 
This formulation allows us to reproduce complex effects such as brushed metals and fabrics. 
Further technical details and equations are provided in Appendix~\ref{appendix:technical-details}.

\subsection{Training Methodology}
\label{section:training-methodology}
To stabilize convergence and reduce interference between reflectance components, we adopt a progressive training strategy. Four components are introduced \revise{in a coarse-to-fine order across simple phases defined by a small 
set of iteration thresholds} (see Appendix~\ref{appendix:implementation-details-progressive-training}, Fig.~\ref{fig:progressive_training_100K} and Fig.~\ref{fig:progressive_training_60K} for details). \revise{A single default configuration is used for all scenes, ensuring that the approach remains stable and reproducible.} Concurrently, we refine camera poses and lighting positions throughout training. The camera adjustment module is activated once the shadow term is introduced, while lighting position refinement begins during the specular phase. %(see Appendix~\ref{appendix:implementation-details-joint-optimization} for details)
Experimental results are presented in our ablation study (see Sec.~\ref{section:ablation-study-schedule} and \revise{Appendix~\ref{appendix:additional-ablation-study-components-schedules}}).

\section{Experiments}
\label{section:experiments}

We evaluate our relightable rendering method on both real-captured and synthetic OLAT datasets. This section first introduces the evaluated methods and datasets, followed by quantitative and qualitative comparisons. We then present ablation studies to assess the contribution of individual model components and training strategies. All experiments are conducted on a workstation equipped with an NVIDIA RTX~3090 GPU and an Intel Core~i7-14700K CPU, running Windows~11 Education.

\subsection{Datasets}

The OLAT datasets provide controlled illumination by sequentially activating individual point light sources, and are widely used benchmarks for evaluating relightable view synthesis. To ensure a consistent and challenging setup, test-time lighting directions are excluded from training.

\paragraph{Real Dataset.} We use the seven OLAT-captured dataset provided by NRHints~\citep{zeng_relighting_2023}: \textit{Cat}, \textit{CatSmall}, \textit{CupFabric}, \textit{Fish}, \textit{FurScene}, \textit{Pikachu}, and \textit{Pixiu}. Each scene contains 500–1500 training images and 45–200 test views, all rendered against a black background. \textit{CatSmall}, \textit{CupFabric}, and \textit{Pikachu} are rendered at a resolution of 1024$\times$1024, while the remaining four use 512$\times$512. 

\paragraph{Synthetic Datasets.} We use the six synthetic scenes released by GS$^3$~\citep{bi_gs3_2024}: \textit{Translucent}, \textit{AnisoMetal}, \textit{Drums}, \textit{FurBall}, \textit{Hotdog}, and \textit{Lego}. Each scene includes 2000 training images and 400 testing images at a resolution of 512$\times$512, rendered against a white background. %The GS$^3$ synthetic dataset is provided in HDR format, and all images are converted to sRGB for fair comparison, as some relighting methods (3DGS~\citep{kerbl_3d_2023}, GI-GS~\citep{chen_gi-gs_2024}, and RNG~\citep{fan_rng_2025}) do not support HDR inputs.
In addition, we evaluate on the five synthetic scenes from SSS-GS~\citep{dihlmann_subsurface_2025}: \textit{Bunny}, \textit{Candle}, \textit{Dragon}, \textit{Soap}, and \textit{Statue}, which emphasize subsurface scattering effects. Each scene includes 500 training images and 500 test views at a downscaled resolution of 256$\times$256 against a black background.

\begin{table*}[t]
    \centering
    \caption{Quantitative comparison results. The best/second-best results are colored in \colorbox[rgb]{1,0.8,0.8}{\strut red} / \colorbox[rgb]{1,0.9,0.8}{\strut orange}.} 
    \vspace{-0.1in}
    \label{tab:quantitative_comparison}
    
	\begin{subtable}{\linewidth}
		\centering
		\caption{Comparison with the original 3DGS~\citep{kerbl_3d_2023}, GI-GS~\citep{chen_gi-gs_2024}, GS$^3$~\citep{bi_gs3_2024}, and RNG~\citep{fan_rng_2025} on the real datasets from NRHints~\citep{zeng_relighting_2023}.}
        \label{tab:quantitative_real}
        \resizebox{\linewidth}{!}{
		\begin{tabular}{l|cc|cc|cc|cc|cc|cc|cc}
			\hline
			\multirow{2}{*}{\diagbox{Method}{Dataset}} & \multicolumn{2}{c|}{Cat} & \multicolumn{2}{c|}{CatSmall} & \multicolumn{2}{c|}{CupFabric} & \multicolumn{2}{c|}{Fish} & \multicolumn{2}{c|}{FurScene} & \multicolumn{2}{c|}{Pikachu} & \multicolumn{2}{c}{Pixiu} \\
			\cline{2-15}
			& Train & Test & Train & Test & Train & Test & Train & Test & Train & Test & Train & Test & Train & Test \\
			\hline
			\multicolumn{15}{c}{\textbf{PSNR} $\uparrow$} \\
			\hline
			3DGS           & 15.2225 & 14.5326 & 22.8367 & 22.5727 & 24.9219 & 25.0488 & 22.8247 & 22.8411 & 18.7746 & 18.4838 & 19.8235 & 19.6310 & 20.0114 & 18.5501 \\
			GI-GS          & 14.5256 & 13.9988 & 22.3667 & 22.3222 & 24.0188 & 24.3821 & 22.2452 & 22.7500 & 17.9882 & 17.8520 & 19.2010 & 19.1867 & 19.0030 & 18.1064 \\
			GS$^3$  & \cellcolor[rgb]{1,0.9,0.8}30.0755 & \cellcolor[rgb]{1,0.9,0.8}27.4081 & \cellcolor[rgb]{1,0.9,0.8}34.8341 & 34.3136 & 36.5090 & 36.1375 & \cellcolor[rgb]{1,0.9,0.8}31.5265 & \cellcolor[rgb]{1,0.9,0.8}30.7218 & \cellcolor[rgb]{1,0.9,0.8}28.6820 & \cellcolor[rgb]{1,0.9,0.8}28.2228 & 30.0745 & 29.4128 & \cellcolor[rgb]{1,0.9,0.8}30.6831 & \cellcolor[rgb]{1,0.9,0.8}29.7001 \\
			RNG            & 27.7478 & 26.6059 & 34.7398 & \cellcolor[rgb]{1,0.9,0.8}34.3709 & \cellcolor[rgb]{1,0.9,0.8}37.7308 & \cellcolor[rgb]{1,0.9,0.8}37.3219 & 29.1378 & 29.0835 & 27.9967 & 27.6930 & \cellcolor[rgb]{1,0.9,0.8}31.6145 & \cellcolor[rgb]{1,0.9,0.8}31.2646 & 29.8650 & 28.8554 \\
			Ours           & \cellcolor[rgb]{1,0.8,0.8}30.0854 & \cellcolor[rgb]{1,0.8,0.8}27.6844 & \cellcolor[rgb]{1,0.8,0.8}35.2740 & \cellcolor[rgb]{1,0.8,0.8}34.6472 & \cellcolor[rgb]{1,0.8,0.8}38.0656 & \cellcolor[rgb]{1,0.8,0.8}37.4702 & \cellcolor[rgb]{1,0.8,0.8}32.0748 & \cellcolor[rgb]{1,0.8,0.8}31.1646 & \cellcolor[rgb]{1,0.8,0.8}31.7846 & \cellcolor[rgb]{1,0.8,0.8}30.7349 & \cellcolor[rgb]{1,0.8,0.8}32.4506 & \cellcolor[rgb]{1,0.8,0.8}31.9298 & \cellcolor[rgb]{1,0.8,0.8}33.6065 & \cellcolor[rgb]{1,0.8,0.8}31.1213 \\
			\hline
			\multicolumn{15}{c}{\textbf{SSIM} $\uparrow$} \\
			\hline
			3DGS           & 0.7140 & 0.6962 & 0.9097 & 0.8896 & 0.9407 & 0.9430 & 0.8424 & 0.8312 & 0.7999 & 0.7869 & 0.9053 & 0.9000 & 0.8624 & 0.8298 \\
			GI-GS          & 0.3210 & 0.3162 & 0.8765 & 0.8750 & 0.9136 & 0.9178 & 0.7430 & 0.7437 & 0.5918 & 0.5811 & 0.8708 & 0.8724 & 0.6229 & 0.6117 \\
			GS$^3$  & \cellcolor[rgb]{1,0.8,0.8}0.9240 & \cellcolor[rgb]{1,0.8,0.8}0.9028 & \cellcolor[rgb]{1,0.9,0.8}0.9777 & \cellcolor[rgb]{1,0.9,0.8}0.9759 & \cellcolor[rgb]{1,0.9,0.8}0.9825 & \cellcolor[rgb]{1,0.9,0.8}0.9821 & \cellcolor[rgb]{1,0.9,0.8}0.9306 & \cellcolor[rgb]{1,0.9,0.8}0.9209 & \cellcolor[rgb]{1,0.9,0.8}0.9426 & \cellcolor[rgb]{1,0.9,0.8}0.9368 & 0.9621 & 0.9605 & \cellcolor[rgb]{1,0.9,0.8}0.9457 & \cellcolor[rgb]{1,0.9,0.8}0.9394 \\
			RNG            & 0.8556 & 0.8427 & 0.9709 & 0.9687 & 0.9803 & 0.9797 & 0.8909 & 0.8923 & 0.9195 & 0.9149 & \cellcolor[rgb]{1,0.9,0.8}0.9673 & \cellcolor[rgb]{1,0.8,0.8}0.9661 & 0.9244 & 0.9187 \\
			Ours           & \cellcolor[rgb]{1,0.9,0.8}0.9224 & \cellcolor[rgb]{1,0.9,0.8}0.9027 & \cellcolor[rgb]{1,0.8,0.8}0.9786 & \cellcolor[rgb]{1,0.8,0.8}0.9767 & \cellcolor[rgb]{1,0.8,0.8}0.9839 & \cellcolor[rgb]{1,0.8,0.8}0.9833 & \cellcolor[rgb]{1,0.8,0.8}0.9363 & \cellcolor[rgb]{1,0.8,0.8}0.9260 & \cellcolor[rgb]{1,0.8,0.8}0.9576 & \cellcolor[rgb]{1,0.8,0.8}0.9518 & \cellcolor[rgb]{1,0.8,0.8}0.9675 & \cellcolor[rgb]{1,0.9,0.8}0.9637 & \cellcolor[rgb]{1,0.8,0.8}0.9524 & \cellcolor[rgb]{1,0.8,0.8}0.9452 \\
			\hline
			\multicolumn{15}{c}{\textbf{LPIPS} $\downarrow$} \\
			\hline
			3DGS           & 0.2983 & 0.3033 & 0.1149 & 0.1183 & 0.0894 & 0.0868 & 0.1700 & 0.1814 & 0.2044 & 0.2101 & 0.1116 & 0.1103 & 0.1473 & 0.1721 \\
			GI-GS          & 0.3496 & 0.3518 & 0.1310 & 0.1339 & 0.1151 & 0.1116 & 0.1996 & 0.2063 & 0.2460 & 0.2492 & 0.1371 & 0.1347 & 0.2957 & 0.3089 \\
			GS$^3$  & \cellcolor[rgb]{1,0.8,0.8}0.1228 & \cellcolor[rgb]{1,0.8,0.8}0.1338 & 0.0624 & 0.0659 & \cellcolor[rgb]{1,0.9,0.8}0.0501 & \cellcolor[rgb]{1,0.9,0.8}0.0506 & \cellcolor[rgb]{1,0.9,0.8}0.0836 & \cellcolor[rgb]{1,0.9,0.8}0.0910 & \cellcolor[rgb]{1,0.9,0.8}0.0778 & \cellcolor[rgb]{1,0.9,0.8}0.0807 & 0.0711 & 0.0717 & \cellcolor[rgb]{1,0.9,0.8}0.0795 & \cellcolor[rgb]{1,0.9,0.8}0.0826 \\
			RNG            & 0.1959 & 0.2023 & \cellcolor[rgb]{1,0.8,0.8}0.0586 & \cellcolor[rgb]{1,0.8,0.8}0.0619 & \cellcolor[rgb]{1,0.8,0.8}0.0373 & \cellcolor[rgb]{1,0.8,0.8}0.0376 & 0.1269 & 0.1292 & 0.1084 & 0.1105 & \cellcolor[rgb]{1,0.8,0.8}0.0507 & \cellcolor[rgb]{1,0.8,0.8}0.0514 & 0.1021 & 0.1057 \\
			Ours           & \cellcolor[rgb]{1,0.9,0.8}0.1251 & \cellcolor[rgb]{1,0.9,0.8}0.1357 & \cellcolor[rgb]{1,0.9,0.8}0.0621 & \cellcolor[rgb]{1,0.9,0.8}0.0656 & 0.0503 & \cellcolor[rgb]{1,0.9,0.8}0.0506 & \cellcolor[rgb]{1,0.8,0.8}0.0779 & \cellcolor[rgb]{1,0.8,0.8}0.0855 & \cellcolor[rgb]{1,0.8,0.8}0.0690 & \cellcolor[rgb]{1,0.8,0.8}0.0724 & \cellcolor[rgb]{1,0.9,0.8}0.0679 & \cellcolor[rgb]{1,0.9,0.8}0.0679 & \cellcolor[rgb]{1,0.8,0.8}0.0751 & \cellcolor[rgb]{1,0.8,0.8}0.0791 \\
			\hline
		\end{tabular}
        }
	\end{subtable}
    
    \vspace{0.5em}
 
    \begin{subtable}{\linewidth}
        \centering
        \caption{Comparison with 3DGS~\citep{kerbl_3d_2023}, GI-GS~\citep{chen_gi-gs_2024}, GS$^3$~\citep{bi_gs3_2024}, and RNG~\citep{fan_rng_2025} on the GS$^3$ synthetic datasets.}
        \label{tab:quantitative_synthetic}
        \resizebox{\linewidth}{!}{
		\begin{tabular}{l|cc|cc|cc|cc|cc|cc}
			\hline
			\multirow{2}{*}{\diagbox{Method}{Dataset}} & \multicolumn{2}{c|}{Translucent} & \multicolumn{2}{c|}{AnisoMetal} & \multicolumn{2}{c|}{Drums} & \multicolumn{2}{c|}{FurBall} & \multicolumn{2}{c|}{Hotdog} & \multicolumn{2}{c}{Lego} \\
			\cline{2-13}
			& Train & Test & Train & Test & Train & Test & Train & Test & Train & Test & Train & Test \\
			\hline
			\multicolumn{13}{c}{\textbf{PSNR} $\uparrow$} \\
			\hline
			3DGS & 17.1853 & 16.4899 & 18.1692 & 17.1009 & 26.5180 & 24.5093 & 21.5473 & 20.1206 & 19.3050 & 16.9535 & 19.0612 & 15.9886 \\
			GI-GS          & 17.1222 & 16.0766 & 17.7309 & 15.9567 & 26.7554 & 24.6177 & 21.3335 & 19.5295 & 19.1535 & 16.8118 & 19.5919 & 16.4229 \\
			GS$^3$ & \cellcolor[rgb]{1,0.9,0.8}31.1327 & \cellcolor[rgb]{1,0.9,0.8}32.1999 & \cellcolor[rgb]{1,0.9,0.8}30.1878 & \cellcolor[rgb]{1,0.9,0.8}28.8219 & \cellcolor[rgb]{1,0.9,0.8}34.0111 & \cellcolor[rgb]{1,0.9,0.8}33.2688 & \cellcolor[rgb]{1,0.9,0.8}34.6201 & \cellcolor[rgb]{1,0.9,0.8}34.9845 & \cellcolor[rgb]{1,0.9,0.8}32.1779 & \cellcolor[rgb]{1,0.8,0.8}32.7244 & \cellcolor[rgb]{1,0.8,0.8}31.2224 & \cellcolor[rgb]{1,0.8,0.8}30.5617 \\
			RNG & 28.1919 & 28.5659 & 26.4611 & 25.9203 & 20.4970 & 20.3033 & 24.5084 & 23.4342 & 29.4095 & 29.5277 & 18.5810 & 18.4872 \\
			Ours & \cellcolor[rgb]{1,0.8,0.8}32.6058 & \cellcolor[rgb]{1,0.8,0.8}32.3919 & \cellcolor[rgb]{1,0.8,0.8}31.1077 & \cellcolor[rgb]{1,0.8,0.8}30.0448 & \cellcolor[rgb]{1,0.8,0.8}34.2448 & \cellcolor[rgb]{1,0.8,0.8}33.5514 & \cellcolor[rgb]{1,0.8,0.8}35.4793 & \cellcolor[rgb]{1,0.8,0.8}35.1639 & \cellcolor[rgb]{1,0.8,0.8}32.4901 & \cellcolor[rgb]{1,0.9,0.8}32.1330 & \cellcolor[rgb]{1,0.9,0.8}31.1434 & \cellcolor[rgb]{1,0.9,0.8}30.4664 \\
			\hline
			\multicolumn{13}{c}{\textbf{SSIM} $\uparrow$} \\
			\hline
			3DGS & 0.8984 & 0.8958 & 0.8995 & 0.8849 & 0.9556 & 0.9439 & 0.9095 & 0.8951 & 0.8956 & 0.8599 & 0.8514 & 0.7904 \\
			GI-GS          & 0.8651 & 0.8586 & 0.8537 & 0.8304 & 0.9066 & 0.8941 & 0.8720 & 0.8592 & 0.8636 & 0.8282 & 0.8128 & 0.7647 \\
			GS$^3$ & \cellcolor[rgb]{1,0.9,0.8}0.9787 & \cellcolor[rgb]{1,0.9,0.8}0.9775 & \cellcolor[rgb]{1,0.9,0.8}0.9702 & \cellcolor[rgb]{1,0.9,0.8}0.9635 & \cellcolor[rgb]{1,0.9,0.8}0.9865 & \cellcolor[rgb]{1,0.9,0.8}0.9841 & \cellcolor[rgb]{1,0.9,0.8}0.9747 & \cellcolor[rgb]{1,0.9,0.8}0.9707 & \cellcolor[rgb]{1,0.9,0.8}0.9764 & \cellcolor[rgb]{1,0.8,0.8}0.9745 & \cellcolor[rgb]{1,0.9,0.8}0.9704 & \cellcolor[rgb]{1,0.8,0.8}0.9581 \\
			RNG & 0.9586 & 0.9598 & 0.9440 & 0.9393 & 0.9199 & 0.9244 & 0.9277 & 0.9204 & 0.9608 & 0.9572 & 0.8756 & 0.8616 \\
			Ours & \cellcolor[rgb]{1,0.8,0.8}0.9835 & \cellcolor[rgb]{1,0.8,0.8}0.9823 & \cellcolor[rgb]{1,0.8,0.8}0.9762 & \cellcolor[rgb]{1,0.8,0.8}0.9698 & \cellcolor[rgb]{1,0.8,0.8}0.9870 & \cellcolor[rgb]{1,0.8,0.8}0.9848 & \cellcolor[rgb]{1,0.8,0.8}0.9776 & \cellcolor[rgb]{1,0.8,0.8}0.9733 & \cellcolor[rgb]{1,0.8,0.8}0.9776 & \cellcolor[rgb]{1,0.9,0.8}0.9743 & \cellcolor[rgb]{1,0.8,0.8}0.9706 & \cellcolor[rgb]{1,0.9,0.8}0.9570 \\
			\hline
			\multicolumn{13}{c}{\textbf{LPIPS} $\downarrow$} \\
			\hline
			3DGS & 0.0755 & 0.0748 & 0.0638 & 0.0704 & 0.0371 & 0.0442 & 0.0918 & 0.0865 & 0.0882 & 0.1128 & 0.1101 & 0.1416 \\
			GI-GS          & 0.1137 & 0.1155 & 0.1084 & 0.1179 & 0.1142 & 0.1242 & 0.1643 & 0.1694 & 0.1368 & 0.1638 & 0.1458 & 0.1643 \\
			GS$^3$ & \cellcolor[rgb]{1,0.9,0.8}0.0247 & \cellcolor[rgb]{1,0.9,0.8}0.0254 & \cellcolor[rgb]{1,0.9,0.8}0.0304 & \cellcolor[rgb]{1,0.9,0.8}0.0341 & \cellcolor[rgb]{1,0.9,0.8}0.0145 & \cellcolor[rgb]{1,0.9,0.8}0.0160 & \cellcolor[rgb]{1,0.9,0.8}0.0566 & \cellcolor[rgb]{1,0.9,0.8}0.0524 & \cellcolor[rgb]{1,0.9,0.8}0.0297 & \cellcolor[rgb]{1,0.8,0.8}0.0305 & \cellcolor[rgb]{1,0.8,0.8}0.0323 & \cellcolor[rgb]{1,0.8,0.8}0.0401 \\
			RNG & 0.0438 & 0.0402 & 0.0490 & 0.0491 & 0.0691 & 0.0685 & 0.1290 & 0.1264 & 0.0473 & 0.0484 & 0.1419 & 0.1470 \\
			Ours & \cellcolor[rgb]{1,0.8,0.8}0.0201 & \cellcolor[rgb]{1,0.8,0.8}0.0200 & \cellcolor[rgb]{1,0.8,0.8}0.0255 & \cellcolor[rgb]{1,0.8,0.8}0.0295 & \cellcolor[rgb]{1,0.8,0.8}0.0144 & \cellcolor[rgb]{1,0.8,0.8}0.0157 & \cellcolor[rgb]{1,0.8,0.8}0.0482 & \cellcolor[rgb]{1,0.8,0.8}0.0442 & \cellcolor[rgb]{1,0.8,0.8}0.0293 & \cellcolor[rgb]{1,0.9,0.8}0.0326 & \cellcolor[rgb]{1,0.9,0.8}0.0331 & \cellcolor[rgb]{1,0.9,0.8}0.0419 \\
			\hline
		\end{tabular}
        }
    \end{subtable}    
    \vspace{0.5em}
    
    \begin{subtable}{\linewidth}
		\centering
		\caption{Comparison with SSS-GS~\citep{dihlmann_subsurface_2025} and KiloOSF~\citep{yu_learning_2022} on the SSS-GS synthetic datasets. For baselines, we report the average results directly from the respective papers, while the per-scene results of our method are provided in the Tab.~\ref{tab:quantitative_synthetic_sss-gs}}
		\label{tab:quantitative_synthetic_sss-gs_overall}
        \resizebox{\linewidth}{!}{
		\begin{tabular}{l|ccc|ccc|ccc}
			\hline
			\multirow{2}{*}{\diagbox{Method}{Dataset}} & \multicolumn{3}{c|}{Train (Average)} & \multicolumn{3}{c|}{Test (Average)} & \multicolumn{3}{c}{Other Metrics} \\
			\cline{2-10}
			& \textbf{PSNR} $\uparrow$ & \textbf{SSIM} $\uparrow$ & \textbf{LPIPS} $\downarrow$ & \textbf{PSNR} $\uparrow$ & \textbf{SSIM} $\uparrow$ & \textbf{LPIPS} $\downarrow$ & \textbf{FPS} & \textbf{Train T.} & \textbf{GPU} \\
			\hline
			KiloOSF & - & - & - & $25.91 \pm 1.88$ & $0.93 \pm 0.02$ & $0.097 \pm 0.03$ & 14.4 & $>20\,\text{h}$ & RTX 4090 \\
			SSS-GS & - & - & - & $35.01 \pm 1.01$ & $0.972 \pm 0.01$ & $0.040 \pm 0.01$ & $154.8 \pm 28.26$ & $<1\,\text{h}$ & RTX 4090 \\
			Ours (w/o Opt) & 40.7087 & 0.9907 & 0.0123 & \cellcolor[rgb]{1,0.9,0.8}37.4409 & \cellcolor[rgb]{1,0.9,0.8}0.9843 & \cellcolor[rgb]{1,0.9,0.8}0.0186 & $66.28 \pm 14.37$ & $<2\,\text{h}$ & RTX 3090 \\
			Ours (w/ Opt) & 41.8705 & 0.9924 & 0.0099 & \cellcolor[rgb]{1,0.8,0.8}38.3542 & \cellcolor[rgb]{1,0.8,0.8}0.9863 & \cellcolor[rgb]{1,0.8,0.8}0.0158 & $61.50 \pm 16.23$ & $\approx 2.5\,\text{h}$ & RTX 3090 \\
			\hline
		\end{tabular}
        }
    \end{subtable} 
    
    \vspace{-1em}
    
\end{table*}

\subsection{Quantitative and Qualitative Analysis}
First, we evaluate both reconstruction quality on the training set and relighting performance on the test set under unseen lighting conditions. %Here, \textit{relighting} refers to inference rendering that produces consistent appearance under novel illumination directions not used during training.

We then compare our method against four representative Gaussian Splatting–based approaches: vanilla 3DGS (baseline), GI-GS~\citep{chen_gi-gs_2024} as a representative of relighting under static illumination, and GS$^3$~\citep{bi_gs3_2024} and RNG~\citep{fan_rng_2025} as representatives of OLAT-based relighting. All methods are trained for 100K iterations with identical settings, and experiments are conducted on both the NRHints real dataset and the GS$^3$ synthetic dataset for fair comparison.

Finally, to further validate the effectiveness of our physically based SSS shading term, we compare against SSS-GS~\citep{dihlmann_subsurface_2025} and KiloOSF~\citep{yu_learning_2022} on the SSS-GS synthetic dataset, using the quantitative results reported in the SSS-GS paper. Following the experimental setup in~\citep{dihlmann_subsurface_2025}, our method is trained for 60K iterations and rendered on a black background to ensure comparability.

\paragraph{Quantitative Results.}
As shown in Tab.~\ref{tab:quantitative_comparison}, our method achieves consistently strong performance across both training and test sets. By introducing a physically based decomposition of shading terms, our approach yields clear numerical advantages on datasets with pronounced scattering and specular effects, while achieving comparable results to other relighting methods on datasets dominated by low-frequency appearance, demonstrating strong generalization across diverse scenarios.

\begin{figure*}[h]
    \centering
    \includegraphics[width=1\linewidth]{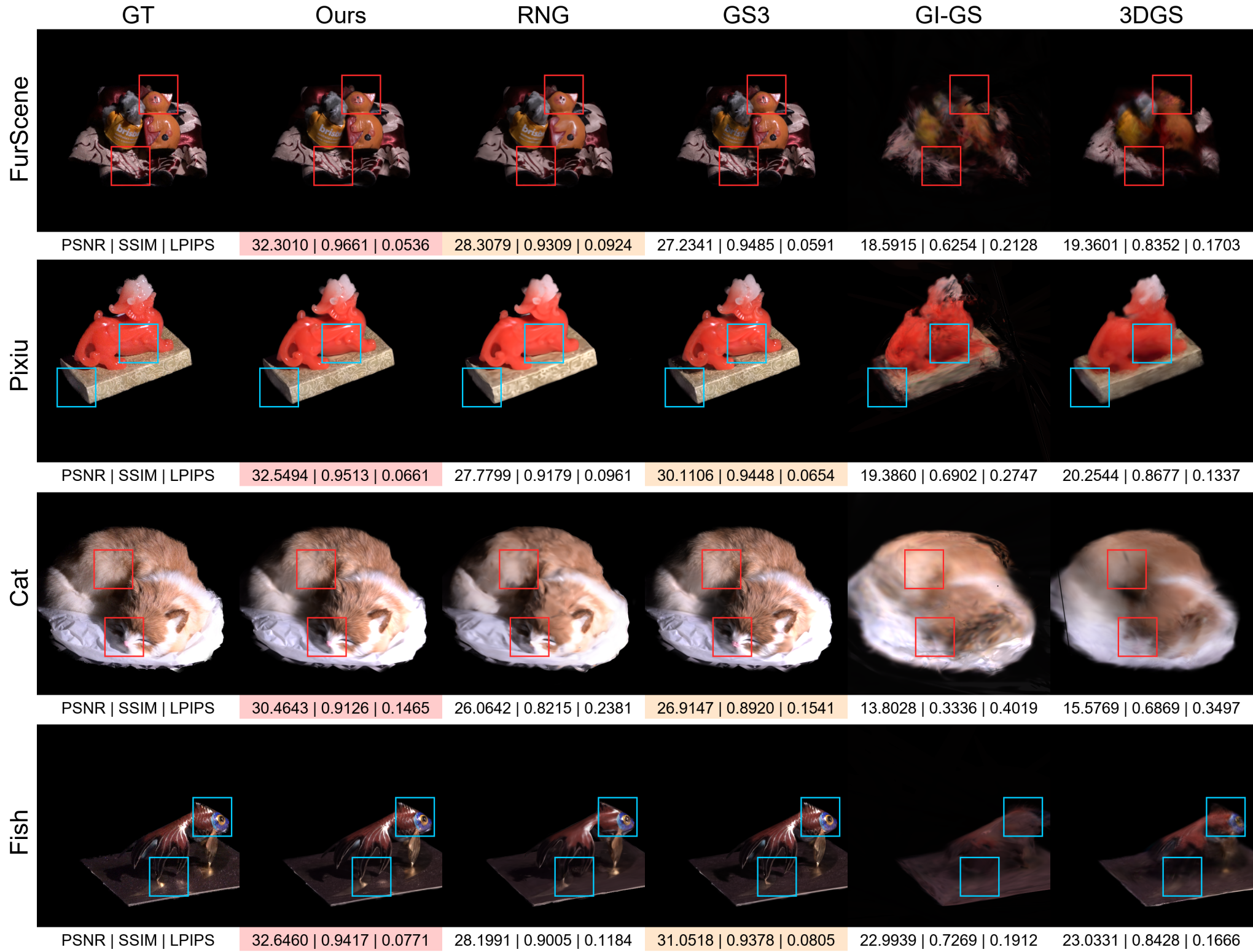}
    \caption{Qualitative comparison on real datasets from NRHints~\citep{zeng_relighting_2023}. It presents relighting results on the test set under novel lighting. The best/second-best results (based on PSNR) are highlighted in \colorbox[rgb]{1,0.8,0.8}{\strut red}/\colorbox[rgb]{1,0.9,0.8}{\strut orange}.}
    \vspace{-0.2in}
    \label{fig:comparison_in_real}
\end{figure*}

\paragraph{Qualitative Results.}
Fig.~\ref{fig:comparison_in_real} presents visual comparisons on the real-world scenes, while additional results on synthetic datasets are provided in Appendix ~\ref{appendix:additional-results-GS3-synthetic-dataset} and~\ref{appendix:additional-results-SSS-GS-synthetic-dataset}.
Compared to existing approaches, our method produces relighting results that are consistently more faithful to the ground truth, especially in challenging scenes with complex material properties and light–material interactions.  
In particular, GS$^3$ often fails to capture sharp shadow boundaries and tends to introduce noise in shadow regions, notably in scenes such as \textit{Fish} and \textit{FurBall}.  
RNG, while capturing reasonable global appearance, frequently loses fine-scale reflectance and geometry details. For instance, the cat’s nose is reconstructed as a flat white region instead of retaining its pink tone and curvature, and specular floor textures in the \textit{Fish} scene are oversmoothed under strong lighting.
These qualitative differences demonstrate our model’s ability to preserve both soft shading and high-frequency details and its robust generalization to unseen lighting.

\subsection{Ablation Study}
\label{section:ablation-study}
%We ablate our components and training strategy to evaluate their effect.

\paragraph{Reflectance Components.}
\label{section:ablation-study-components}
We evaluate different combinations of reflectance terms to understand their individual and cumulative contributions. Specifically, we compare: (A) Diffuse only, serving as a baseline; (B) adding specular; (C) adding subsurface scattering; and (D) the full model with all terms. We also examine ablations from the full model by removing: (E) the specular term, or (F) the scattering term. The full model (D) achieves the best overall performance. Subtractive ablations confirm these trends: removing either specular (E) or scattering (F) leads to noticeable degradation.

\paragraph{Training Schedule.}
\label{section:ablation-study-schedule}
We examine alternative strategies for introducing reflectance components during training: (H) joint training of all terms from the start; (I) our progressive schedule (Diffuse → Shadow → Scatter → Specular); (J) a non-physical variant swapping the last two; and (K) a variant that adds all terms together after a diffuse-only warm-up. The results demonstrate the effectiveness of our progressive strategy (I), which yields superior reconstruction of reflectance components.
\begin{figure*}[t]
    \centering
    \includegraphics[width=1\linewidth]    {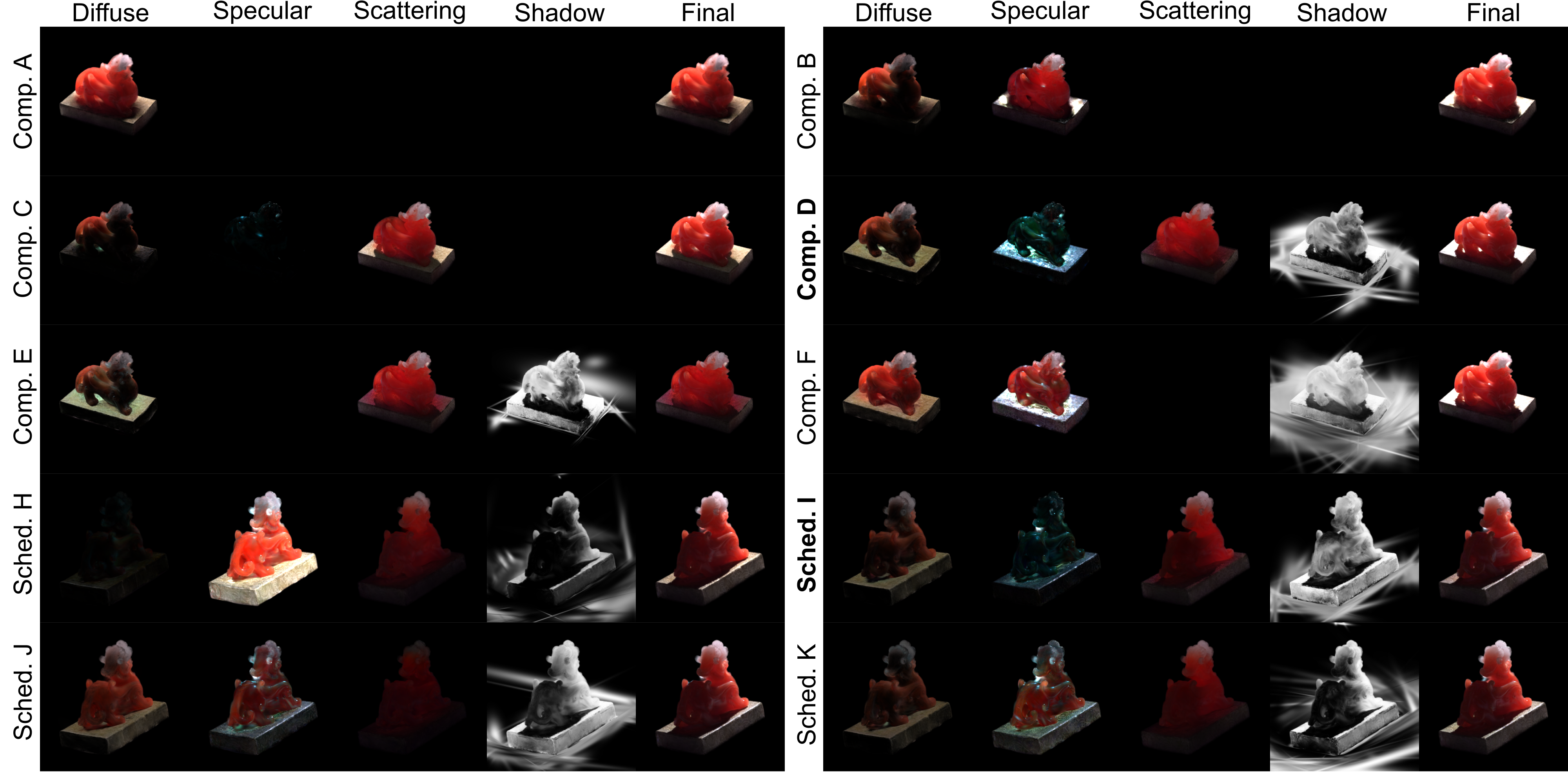}
    \vspace{-0.2in}
    \caption{Visualization of reconstructed components under different reflectance decompositions and training schedules on the \textit{Pixiu} scene from the 
NRHints real dataset. Top: six reflectance compositions (Comp. A-F); Bottom: four training schedules (Sched. H-K).}
    \label{fig:ablationstudy_pixiu}
\end{figure*}

\begin{table}[t]
    \centering
    \caption{Ablation study results on the real scene \textit{Pixiu}.}
    \vspace{-0.1in}
    \label{tab:ablation_study}

    \resizebox{0.7\linewidth}{!}{
    \begin{tabular}{c|ccc|ccc}
        \hline
        \multicolumn{1}{c|}{\multirow{2}{*}{\diagbox{Method}{Dataset}}} 
        & \multicolumn{3}{c|}{Train Set} & \multicolumn{3}{c|}{Test Set} \\
        \cline{2-7}
        & PSNR $\uparrow$ & SSIM $\uparrow$ & LPIPS $\downarrow$ 
        & PSNR $\uparrow$ & SSIM $\uparrow$ & LPIPS $\downarrow$ \\
        \hline
        A: Diff & 20.2869 & 0.5701 & 0.1055 & 20.1878 & 0.5583 & 0.1061 \\
        B: D + S & 20.7321 & 0.6800 & 0.0966 & 20.5692 & 0.6683 & 0.0992 \\
        C: D + S + SSS & 25.1545 & 0.9336 & 0.0825 & 24.8100 & 0.9274 & 0.0857 \\
        D: Full (Ours) & \cellcolor[rgb]{1,0.8,0.8}33.6065 & \cellcolor[rgb]{1,0.8,0.8}0.9524 & \cellcolor[rgb]{1,0.8,0.8}0.0751 & \cellcolor[rgb]{1,0.8,0.8}31.1213 & \cellcolor[rgb]{1,0.8,0.8}0.9452 & \cellcolor[rgb]{1,0.8,0.8}0.0791 \\
        E: Full – S & \cellcolor[rgb]{1,0.9,0.8}32.3487 & \cellcolor[rgb]{1,0.9,0.8}0.9489 & \cellcolor[rgb]{1,0.9,0.8}0.0817 & \cellcolor[rgb]{1,0.9,0.8}30.5952 & \cellcolor[rgb]{1,0.9,0.8}0.9429 & \cellcolor[rgb]{1,0.9,0.8}0.0844 \\
        F: Full – SSS & 31.5332 & 0.7318 & 0.0818 & 30.4095 & 0.7204 & 0.0850 \\
        \hline
		H: Joint & 32.5452 & 0.9500 & 0.0781 & 31.0880 & 0.9441 & 0.0812 \\
		I: Prog. Phys (Ours) & \cellcolor[rgb]{1,0.8,0.8}33.6065 & \cellcolor[rgb]{1,0.8,0.8}0.9524 & \cellcolor[rgb]{1,0.8,0.8}0.0751 & \cellcolor[rgb]{1,0.8,0.8}31.1213 & \cellcolor[rgb]{1,0.8,0.8}0.9452 & \cellcolor[rgb]{1,0.8,0.8}0.0791 \\
		J: Prog. NonPhys &	32.5606 & 0.9499 & 0.0776 & \cellcolor[rgb]{1,0.9,0.8}31.0973 & 0.9443 & 0.0807 \\
		K: Prog. Merge & \cellcolor[rgb]{1,0.9,0.8}33.3438 & \cellcolor[rgb]{1,0.9,0.8}0.9520 & \cellcolor[rgb]{1,0.9,0.8}0.0758 & 31.0486 & \cellcolor[rgb]{1,0.9,0.8}0.9449 & \cellcolor[rgb]{1,0.9,0.8}0.0794 \\
        \hline
    \end{tabular}
    }
\end{table}

While the quantitative results across different compositions and training schedules remain relatively close (see Tab.~\ref{tab:ablation_study} \revise{and the additional scenes in Tab.~\ref{tab:ablation_study_additional-scene}}), the visual decompositions (see Fig.~\ref{fig:ablationstudy_pixiu} \revise{and Fig.~\ref{fig:ablationstudy_translucent}}) show meaningful differences that highlight the importance of proper terms and training strategies. 
For example, in \textit{Composition F}, removing the scattering term leads to noticeable artifacts, where both the diffuse and shadow components begin to absorb scattering, resulting in a more translucent appearance that compromises the sharpness of shadows. 
Similarly, in \textit{Schedule K}, introducing multiple reflectance terms simultaneously causes training interference, where overlapping gradients between specular and scattering degrade the disentanglement quality. 
These artifacts are less evident in scalar metrics but manifest clearly in the visual outputs, underscoring the need for structured supervision and progressive learning.

\section{Conclusion}

We demonstrate that progressively introducing reflectance terms via a carefully designed training schedule enables our method to decompose scene illumination effectively and support relighting under novel lighting. \revise{Although we do not explicitly model multi-bounce global illumination, the combination of continuous volumetric visibility and the learned scattering term already captures the most perceptually important low-frequency indirect effects.} While our current implementation relies on a rasterization-based pipeline, which does not fully capture physical light transport, future work could integrate ray or path tracing to improve physical realism. Incorporating additional supervision, such as multi-term losses, may further reduce role leakage and improve disentanglement of reflectance components. Material-aware grouping of Gaussians using the learned material latent space could produce a more structured representation, facilitating controllable relighting and scene editing. Overall, our work establishes a solid foundation for physically grounded, editable relightable rendering.

\paragraph{Ethics Statement.}
This work introduces a physically based relightable 3D reconstruction framework that recovers geometry and appearance from sparse image observations. The method aims to preserve the visual fidelity of captured scenes and does not infer personal identity or generate content beyond illumination variation. As with other reconstruction techniques, it may be applied to data containing human subjects or proprietary objects, and thus should be used in accordance with relevant consent, privacy, and intellectual property regulations.

\paragraph{Reproducibility Statement.}
We provide sufficient information in the main paper (Sec.~\ref{section:experiments}) and the appendix (Sec.~\ref{appendix:implementation-details} and Sec.~\ref{appendix:technical-details}) to support reproducibility, including details on the model design, training procedure, and evaluation setup. All datasets used for training and testing are described accordingly. We are happy to clarify any additional implementation details if needed.

\paragraph{Acknowledgements.}
This work was supported in part by the Marsden Fund Council managed by the
Royal Society of New Zealand under Grant MFP-20-VUW-180, and internal research grant (Project No. 400876) from Victoria University of Wellington.

\bibliography{all-reference}
\bibliographystyle{iclr2026_conference}

\clearpage

\appendix
\section{Implementation Details}
\label{appendix:implementation-details}
\begin{figure*}[h]
    \centering
    \includegraphics[width=1\linewidth]    {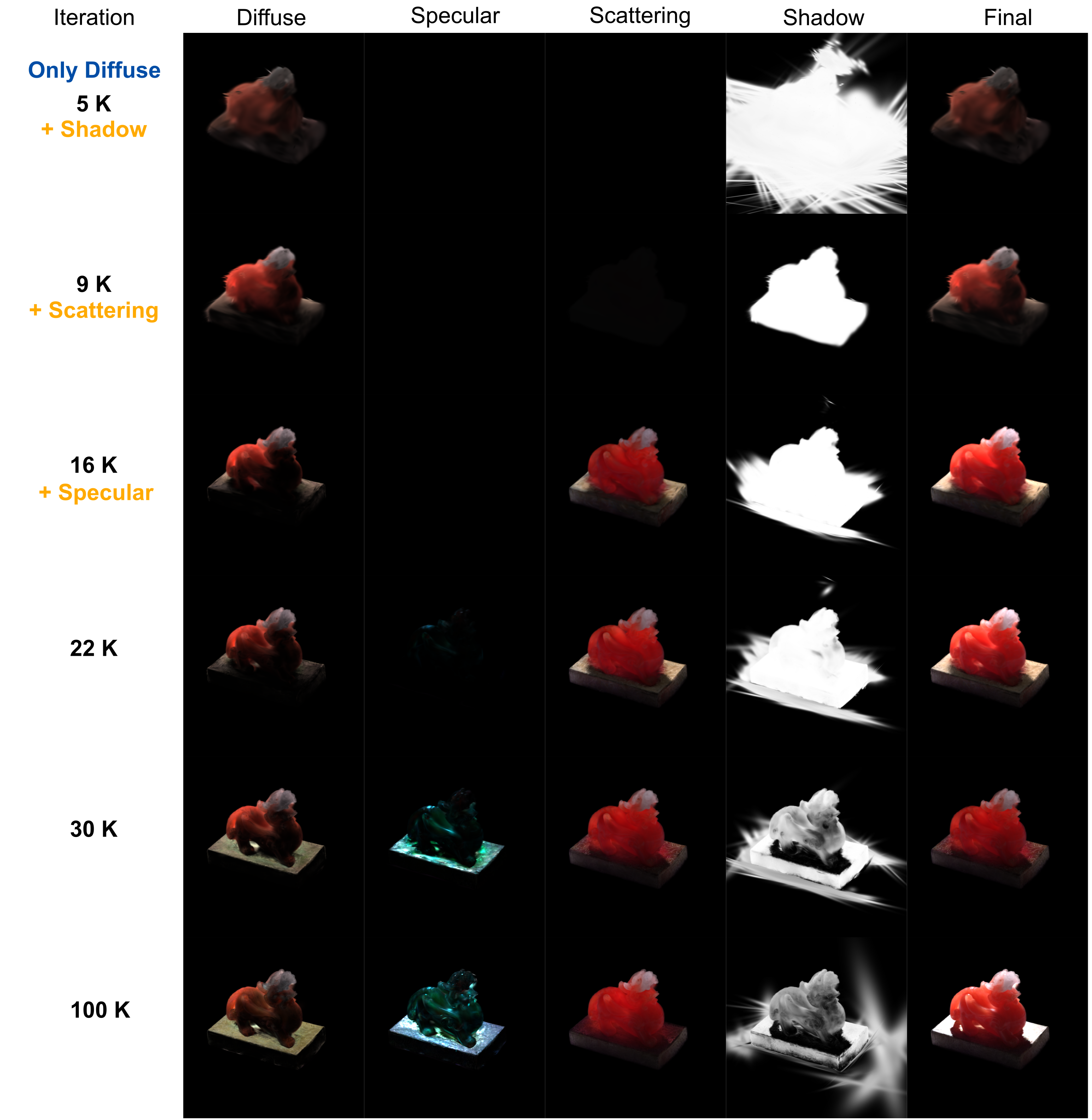}
    \caption{Illustration of the progressive training schedule on the \textit{pixiu} scene from the NRHints real dataset~\citep{zeng_relighting_2023}. The model is trained for a total of 100K iterations. }
    \label{fig:progressive_training_100K}
\end{figure*}

\begin{figure*}[h]
    \centering
    \includegraphics[width=1\linewidth]    {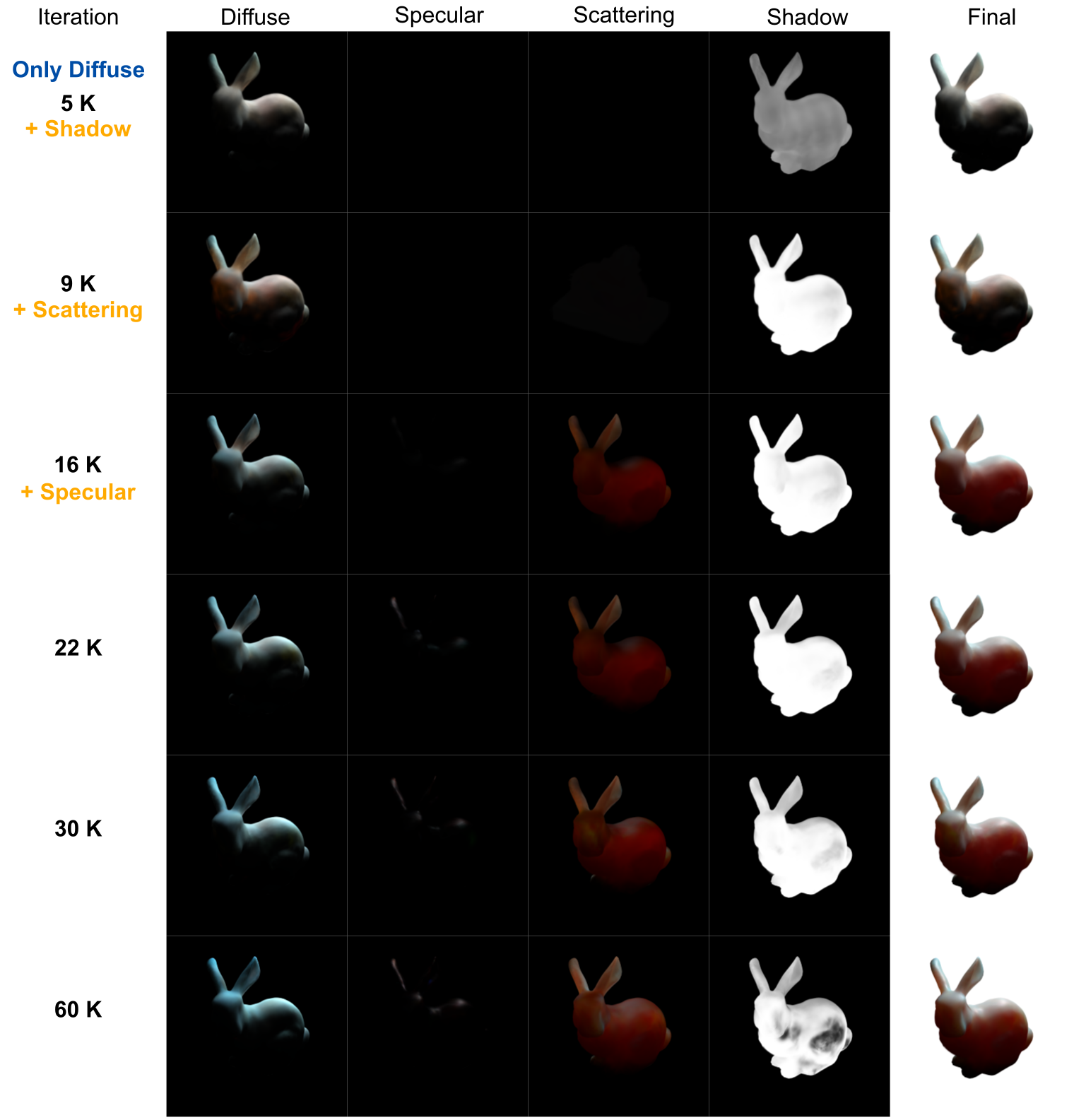}
    \caption{Illustration of the progressive training schedule on the \textit{bunny} scene from the SSS-GS synthetic dataset~\citep{dihlmann_subsurface_2025}. The model is trained for a total of 60K iterations. }
    \label{fig:progressive_training_60K}
\end{figure*}

\refstepcounter{subsection}
\paragraph{Progressive Training.}
\label{appendix:implementation-details-progressive-training}
Our experiments follow the reflectance \textit{Composition D} and the progressive \textit{Schedule I} described in Sec.~\ref{section:ablation-study}. For fair comparison, we train each scene for 100K iterations on the NRHints and GS$^3$ datasets, and for 60K iterations on the SSS-GS dataset to match the original settings. The overall progressive training process is illustrated in Fig.~\ref{fig:progressive_training_100K} and Fig.~\ref{fig:progressive_training_60K}.

Specifically, during the initial 5K iterations, only the diffuse term contributes to shading. This \revise{warm-up} phase stabilizes coarse geometry and appearance, since all Gaussians are randomly initialized (10K points uniformly distributed on the unit sphere) from NeRF JSON inputs. These early diffuse-only iterations are therefore critical for forming a reliable Gaussian structure.

From 5K iterations onward, we introduce the shadow term, which provides a first approximation of light visibility and substantially improves lighting initialization. At 9K iterations, the subsurface-scattering term is activated to \revise{model low-frequency, multiple-scattering effects in translucent materials}. Between 9K and 16K iterations, shadow \revise{gradients are temporarily held fixed so that} the scattering term can converge without interference from competing gradients. This schedule is motivated by the longer convergence needs of scattering, which would otherwise remain underfitted.

After 16K iterations, we introduce the specular term to capture high-frequency details and view-dependent highlights. \revise{Optimizing specular too early tends to dominate smoother reflectance components, especially subsurface scattering, since the optimizer naturally prioritizes sharper signals} (see \revise{Sched. H in Fig.~\ref{fig:ablationstudy_pixiu} and Fig.~\ref{fig:ablationstudy_translucent}}). To address this, we temporarily freeze gradients of the scattering term between 13K and 20K iterations, allowing specular learning to progress without suppressing low-frequency transport. Additionally, from 16K to 20K iterations we suspend updates to the ASG lobe parameters (scale and rotation), so the specular module first focuses on Fresnel intensity before refining anisotropic lobe orientations.

In summary, the progressive schedule gradually disentangles low- and high-frequency reflectance phenomena, balancing the convergence speed of each component and yielding a stable and physically consistent optimization. 

% \refstepcounter{subsection}
% \paragraph{Joint Camera–Light Optimization.}
% \label{appendix:implementation-details-joint-optimization}

\begin{figure}[t]
    \centering
    \includegraphics[width=1\linewidth]{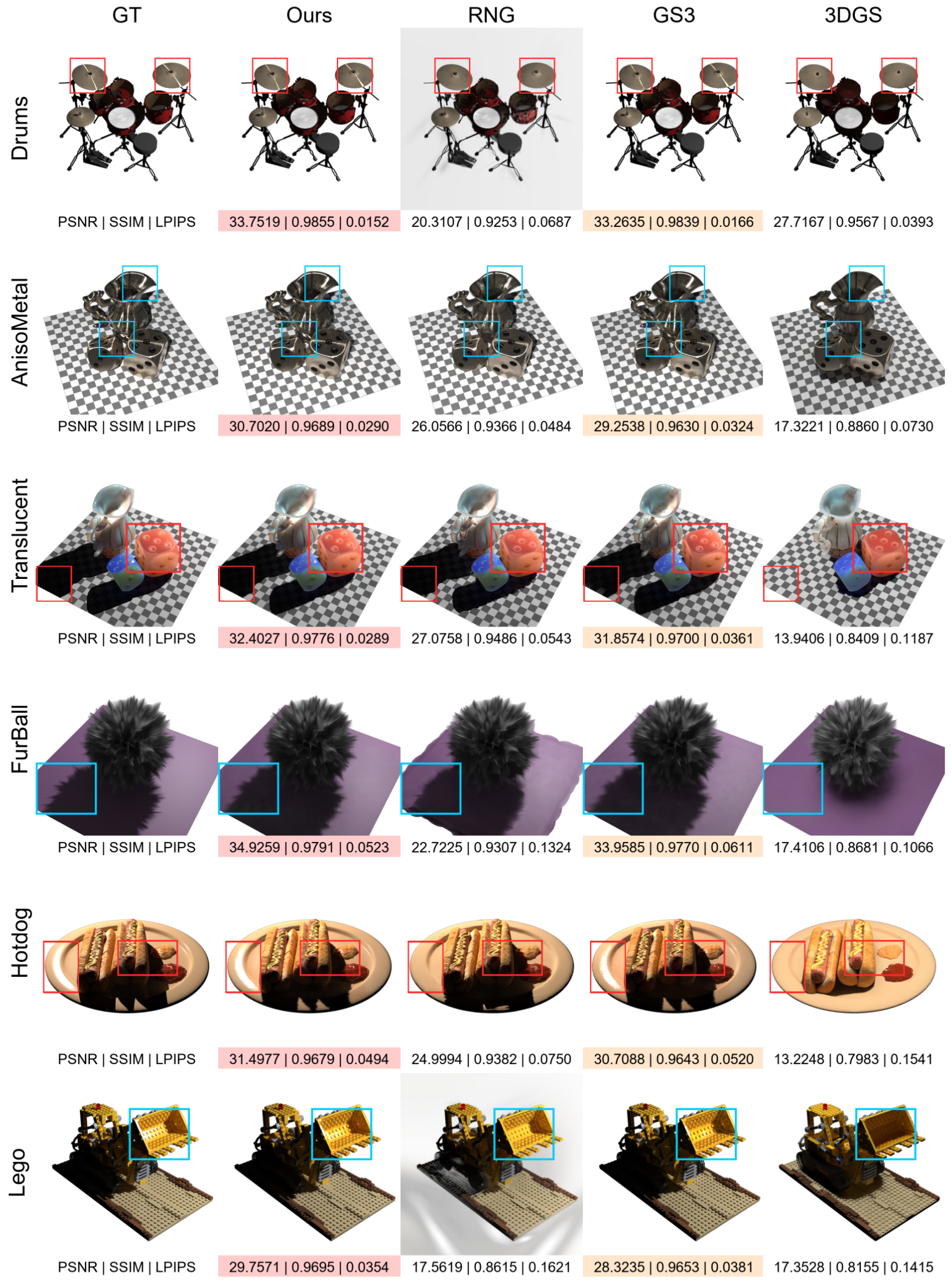}
    \caption{Qualitative comparison of relighting results on novel test-time lighting from synthetic datasets provided by $\text{GS}^3$~\citep{bi_gs3_2024}. Note that the rendered view of RNG's results on the \textit{Drums} and \textit{Lego} scenes are obtained using their official code trained using a set of white-background images.} 
 %The left column shows reconstruction results on training views, while the right column presents relighting results on novel test-time lighting. 
 %The best and second-best results (based on PSNR) are highlighted in \colorbox[rgb]{1,0.8,0.8}{\strut red} and \colorbox[rgb]{1,0.9,0.8}{\strut orange}, respectively.
    \label{fig:comparison_in_syn_test_gs3}
\end{figure}

\section{Technical Details}
\label{appendix:technical-details}
\paragraph{Diffuse Term.}
We adopt a Lambertian BRDF to model the diffuse component, assuming view-independent cosine-weighted reflection:
\begin{equation}
f_d = \max(0, \mathbf{n} \cdot \omega_i),
\label{eq:diffuse}
\end{equation}
where $\mathbf{n}$ is the surface normal and $\omega_i$ is the incident light direction. This simple formulation stabilizes the reconstruction of low-frequency reflectance and provides a physically interpretable baseline for shading.

\paragraph{Specular Term.}
The specular component is modeled as a Fresnel factor multiplied by an anisotropic spherical Gaussian (ASG) reflectance function:
\begin{equation}
f_s = F(\omega_o, \mathbf{h}) \cdot D_{\text{ASG}}(\mathbf{h}),
\label{eq:specular}
\end{equation}
where $\mathbf{h}$ is the half-vector between light and view directions, and $F(\omega_o, \mathbf{h})$ denotes the Fresnel reflectance term, approximated with Schlick’s formulation~\citep{schlick_inexpensive_1994}. 

The ASG reflectance function is expressed as a weighted sum of $N$ global ASG bases:
\begin{equation}
D_{\text{ASG}}(\mathbf{h}) = \sum_{j=1}^{N} G_j(\mathbf{h}) \cdot \alpha_j,
\end{equation}
where each basis $G_j$ takes the geometric form~\citep{xu_anisotropic_2013}:
\begin{equation}
G_{j}(\mathbf{h}) = \exp\left( -\lambda_j (\mathbf{h} \cdot \mathbf{x}_j)^2 - \mu_j (\mathbf{h} \cdot \mathbf{y}_j)^2 \right),
\end{equation}
with $(\mathbf{x}_j, \mathbf{y}_j)$ defining the local anisotropy axes. By leveraging a compact set of global ASG bases with learnable weights $\alpha_j$, we achieve expressive, view-dependent reflections without the need for per-Gaussian specular storage, maintaining efficiency while preserving rendering fidelity.

\begin{table*} [h]
    \centering
    \caption{Quantitative comparison with SSS-GS~\citep{dihlmann_subsurface_2025} and KiloOSF~\citep{yu_learning_2022} on the synthetic datasets provided by SSS-GS, trained for 60K iterations. The best and second-best results (based on PSNR) are highlighted in \colorbox[rgb]{1,0.8,0.8}{\strut red} and \colorbox[rgb]{1,0.9,0.8}{\strut orange}, respectively.}
    \label{tab:quantitative_synthetic_sss-gs}

	\resizebox{\textwidth}{!}{
		\begin{tabular}{l|cc|cc|cc|cc|cc|cc}
			\hline
			\multirow{2}{*}{\diagbox{Method}{Dataset}} & \multicolumn{2}{c|}{Bunny} & \multicolumn{2}{c|}{Candle} & \multicolumn{2}{c|}{Dragon} & \multicolumn{2}{c|}{Soap} & \multicolumn{2}{c|}{Statue} & \multicolumn{2}{c}{Average} \\
			\cline{2-13}
			& Train & Test & Train & Test & Train & Test & Train & Test & Train & Test & Train & Test \\
			\hline
			\multicolumn{13}{c}{\textbf{PSNR} $\uparrow$} \\
			\hline
			KiloOSF & - & - & - & - & - & - & - & - & - & - & - & $25.91 \pm 1.88$ \\
			SSS-GS & - & - & - & - & - & - & - & - & - & - & - & $35.01 \pm 1.01$ \\
			Ours (w/o Opt) & 40.7672 & 37.2270 & 40.0682 & 38.3662 & 39.3646 & 36.6325 & 45.1439 & 40.4385 & 38.1997 & 34.5404 & 40.7087 & \cellcolor[rgb]{1,0.9,0.8}37.4409 \\
			Ours (w/ Opt) & 40.8704 & 37.2960 & 43.2426 & 41.1038 & 40.8462 & 37.3363  & 45.2659 & 40.6914 & 39.1271 & 35.3434 & 41.8705 & \cellcolor[rgb]{1,0.8,0.8}38.3542 \\
			\hline
			\multicolumn{13}{c}{\textbf{SSIM} $\uparrow$} \\
			\hline
			KiloOSF & - & - & - & - & - & - & - & - & - & - & - & $0.93 \pm 0.02$ \\
			SSS-GS & - & - & - & - & - & - & - & - & - & - & - & $0.972 \pm 0.01$ \\
			Ours (w/o Opt) & 0.9922 & 0.9859 & 0.9910 & 0.9875 & 0.9874 & 0.9789 & 0.9950 & 0.9908 & 0.9879 & 0.9782 & 0.9907 & \cellcolor[rgb]{1,0.9,0.8}0.9843 \\
			Ours (w/ Opt) & 0.9920 & 0.9861 & 0.9947 & 0.9921 & 0.9900 & 0.9812 & 0.9951 & 0.9912 & 0.9901 & 0.9812 & 0.9924 & \cellcolor[rgb]{1,0.8,0.8}0.9863 \\
			\hline
			\multicolumn{13}{c}{\textbf{LPIPS} $\downarrow$} \\
			\hline
			KiloOSF & - & - & - & - & - & - & - & - & - & - & - & $0.83 \pm 0.09$ \\
			SSS-GS & - & - & - & - & - & - & - & - & - & - & - & $0.040 \pm 0.01$ \\
			Ours (w/o Opt) & 0.0113 & 0.0179 & 0.0134 & 0.0172 & 0.0157 & 0.0240 & 0.0060 & 0.0104 & 0.0153 & 0.0235 & 0.0123  & \cellcolor[rgb]{1,0.9,0.8}0.0186 \\
			Ours (w/ Opt) & 0.0111 & 0.0173 & 0.0072 & 0.0099 & 0.0121 & 0.0209 & 0.0056 & 0.0099 & 0.0135 & 0.0212 & 0.0099 & \cellcolor[rgb]{1,0.8,0.8}0.0158 \\
			\hline
		\end{tabular}
	}
\end{table*}

\section{Additional Results on the Synthetic Dataset}
\refstepcounter{subsection}
\paragraph{GS$^3$ Synthetic Dataset.}
\label{appendix:additional-results-GS3-synthetic-dataset}
We compare our method against four representative baselines: the original 3D Gaussian Splatting~\citep{kerbl_3d_2023}, GI-GS~\citep{chen_gi-gs_2024}, $\text{GS}^3$~\citep{bi_gs3_2024}, and RNG~\citep{fan_rng_2025}. These methods are evaluated on reconstruction of training views from the synthetic datasets released by $\text{GS}^3$, which contain diverse reflectance properties and serve as a comprehensive benchmark for relightable rendering. Quantitative results are reported in Tab.~\ref{tab:quantitative_synthetic} of the main paper, and qualitative comparisons are shown in Fig.~\ref{fig:comparison_in_syn_test_gs3}. \revise{We additionally compare against several representative relighting baselines, as presented in Appendix~\ref{appendix:additional-relighting-baselines}.}

To ensure a fair comparison, all methods were trained for 100K iterations on the same training sets, followed by rendering on both the training and test splits. Reconstruction quality is evaluated on the training set, while relighting performance is assessed on the test set, which contains unseen lighting conditions. Our method achieves superior reconstruction of specular highlights across most scenes, such as \textit{Drums} and \textit{Hotdog}, and demonstrates more accurate shadow reconstruction in complex cases like \textit{FurBall}.

% Notably, in this experiment, all methods were trained using a white background for consistency. However, the original RNG implementation~\citep{fan_rng_2025} was trained with a black background. Due to time constraints, we did not retrain RNG and the other methods under this setting. While we have not explicitly verified the impact, this background mismatch may partially explain RNG’s relatively lower performance on the synthetic datasets. Its original black-background setting is preserved in the real dataset comparisons, as shown in Tab.~\ref{tab:quantitative_real} and Fig.~\ref{fig:comparison_in_real} of the main paper.

Additionally, we observed that $\text{GS}^3$ occasionally produces shadow artifacts at certain viewpoints—especially in scenes like \textit{FurBall}—and we intentionally avoided including such anomalous views in the comparison to maintain fairness.

Since the $\text{GS}^3$ synthetic datasets contain limited subsurface scattering effects, we further validate the effectiveness of our model on the recently released synthetic dataset from SSS-GS, which features more prominent subsurface scattering phenomena.

\begin{figure}[t]
    \centering
    \includegraphics[width=1\linewidth]{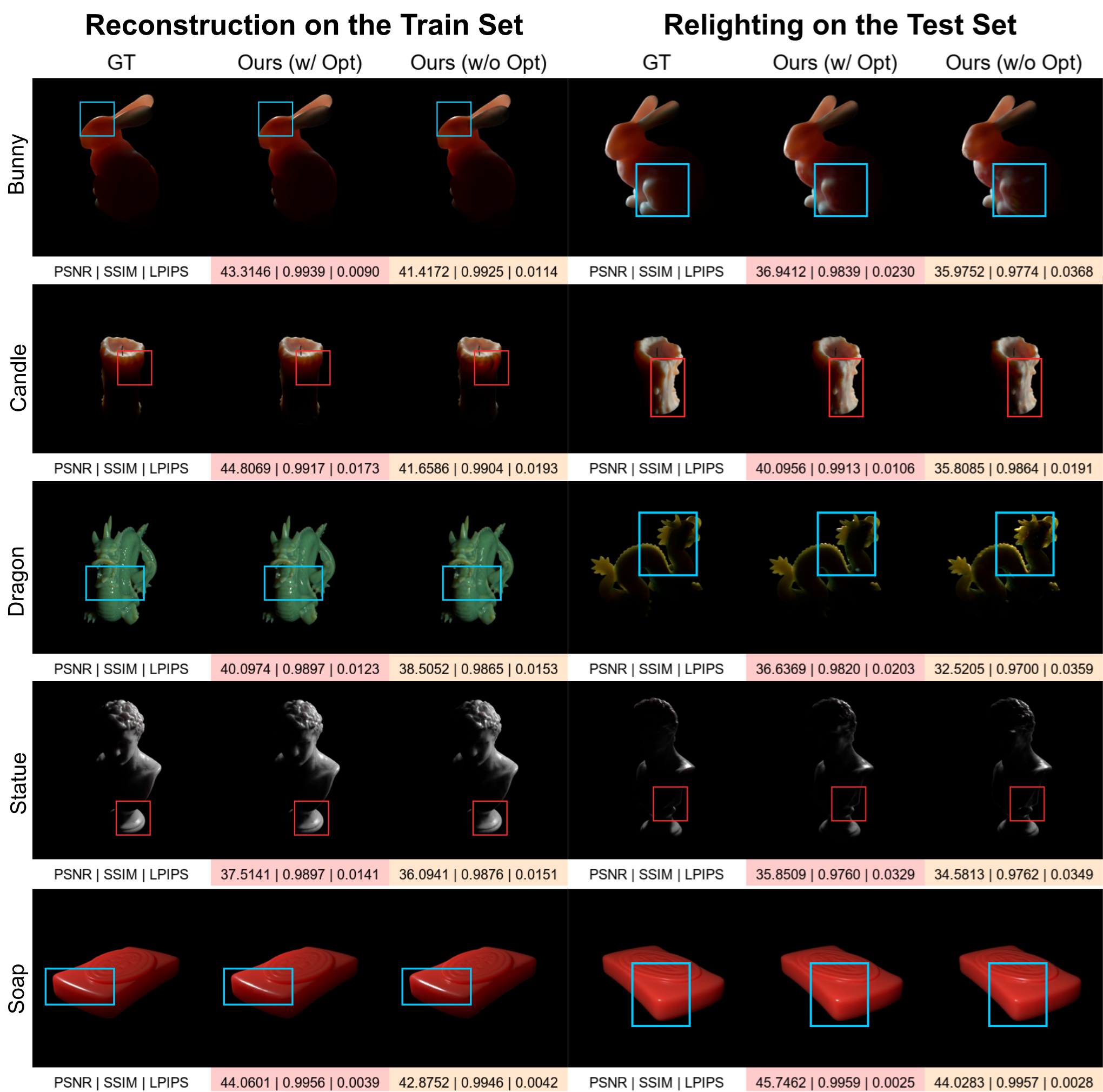}
    \caption{Qualitative comparison on synthetic datasets from SSS-GS~\citep{dihlmann_subsurface_2025}. The left column shows reconstruction results on training views, while the right column presents relighting results on novel test-time lighting. The best and second-best results (based on PSNR) are highlighted in \colorbox[rgb]{1,0.8,0.8}{\strut red} and \colorbox[rgb]{1,0.9,0.8}{\strut orange}, respectively.}
    \label{fig:comparison_in_syn_sss-gs}
\end{figure}

\refstepcounter{subsection}
\paragraph{SSS-GS Synthetic Dataset.}
\label{appendix:additional-results-SSS-GS-synthetic-dataset}
We further evaluate our method on the recently released synthetic dataset from SSS-GS~\citep{dihlmann_subsurface_2025}. While both scale-down and full-resolution versions of the synthetic dataset exist, only the scale-down version (500 images per split at 256$\times$256 resolution) is publicly available. In contrast, the real datasets are currently released only in full resolution (13,193 images per split at 800$\times$649). Due to the limited time, we conducted our experiments only on the publicly available synthetic subset. This includes five scenes: \textit{bunny}, \textit{candle}, \textit{dragon}, \textit{soap}, and \textit{statue}. 

\begin{figure}[t]
    \centering
    \includegraphics[width=1\linewidth]{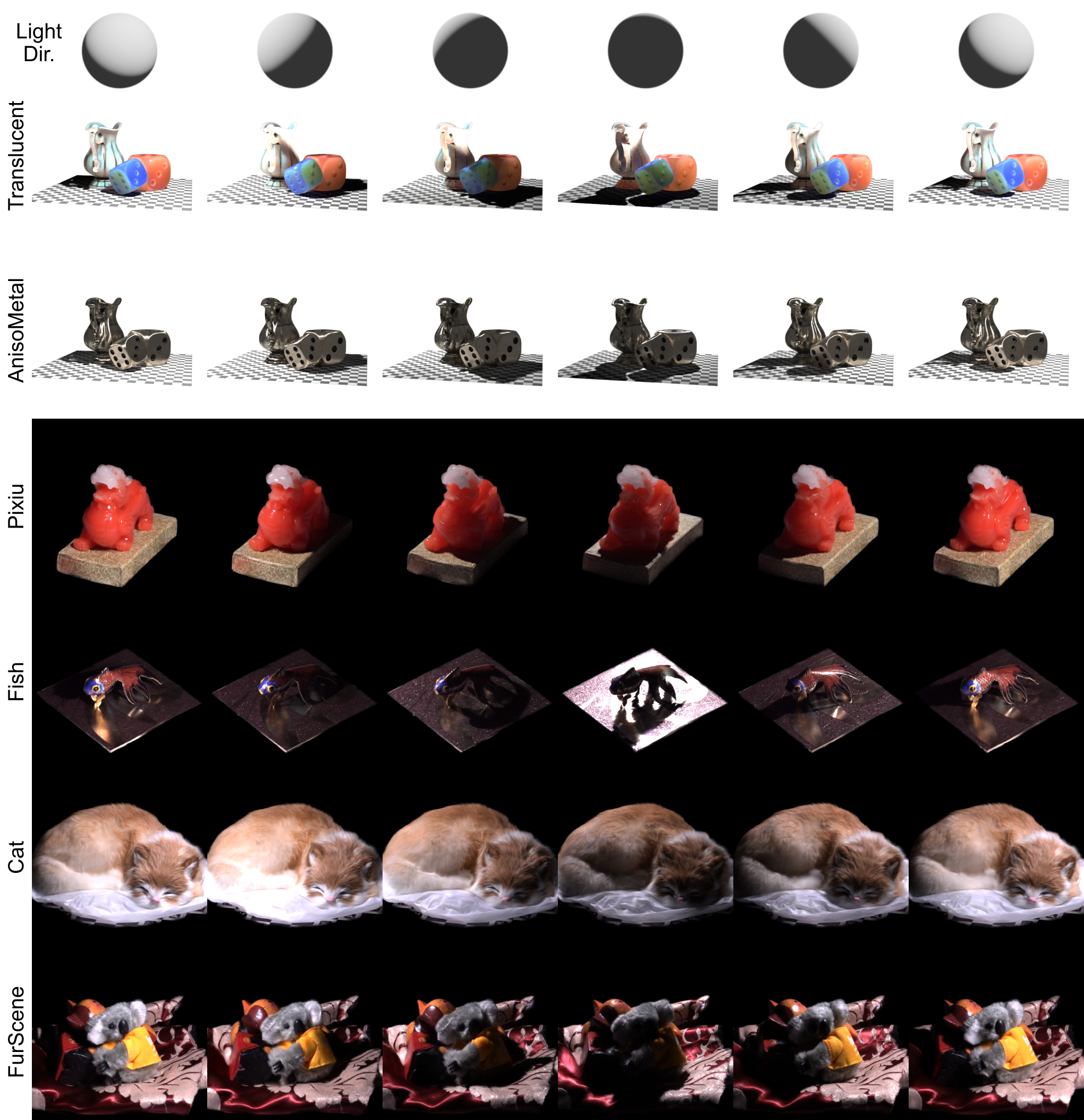}
    \caption{Qualitative relighting results on both real and synthetic datasets. 
Each scene is rendered from the same camera viewpoint under six novel lighting conditions.}
    \label{fig:relighting}
\end{figure}

Following the experimental setup in the SSS-GS paper~\citep{dihlmann_subsurface_2025}, our method was trained for 60K iterations and rendered on a black background to ensure a fair comparison. We compare our method against SSS-GS and KiloOSF~\citep{yu_learning_2022} using the quantitative results reported in the SSS-GS paper. Quantitative results are summarized in Tab.~\ref{tab:quantitative_synthetic_sss-gs_overall}, which reports overall metrics including average reconstruction quality and training cost. Per-scene quantitative results are further detailed in Tab.~\ref{tab:quantitative_synthetic_sss-gs}, allowing a finer-grained comparison across individual scenes. Qualitative comparisons are provided in Fig.~\ref{fig:comparison_in_syn_sss-gs}.

We include two variants of our method in this comparison: Ours (w/o Opt) refers to our approach without camera and lighting optimization, while Ours (w/ Opt) includes joint optimization of both camera poses and light directions. This setting was briefly introduced in Sec.~\ref{section:training-methodology} of the main paper, where we progressively refine camera poses and lighting positions during training. The inclusion of camera and lighting optimization (w/ Opt) leads to noticeable improvements in both camera and light estimates, resulting in an average increase of approximately 1dB in PSNR on both training and test sets. Despite the input images being downscaled to 256$\times$256, subtle lighting differences remain discernible, particularly in the relighting results on test views of scenes such as \textit{bunny} and \textit{dragon} (right columns).

\section{Analysis of Relighting}
To complement the OLAT test set, which provides ground-truth images for quantitative evaluation, we construct a validation setup for relighting with synthetic light–camera trajectories stored in JSON format. In this setting, each scene is illuminated by a single point light source, and the incident light direction is derived from the relative position between the light source and the center of each Gaussian, ensuring shading consistency at the Gaussian level. 
Unlike the test set, the validation setup does not include ground-truth images and is instead used to assess generalization under novel light and view configurations. Specifically, the light source is placed on a circular path around the object, sweeping $360^\circ$ in azimuth with a step of $2.4^\circ$. Next, the camera is rotated by $180^\circ$ to the back side of the object with a step of $2^\circ$, after which the light completes another full $360^\circ$ sweep around the back view. Finally, the camera is rotated $180^\circ$ back to the original frontal view. 

This configuration provides dense sampling of both lighting and viewing conditions, enabling a comprehensive assessment of relighting fidelity. We perform relighting under this setup for both real and synthetic datasets (see Fig.~\ref{fig:relighting}). The results demonstrate that our method maintains consistency across varying illumination while faithfully preserving view-dependent effects.

\begin{figure}[h]
    \centering
    \includegraphics[width=1\linewidth]{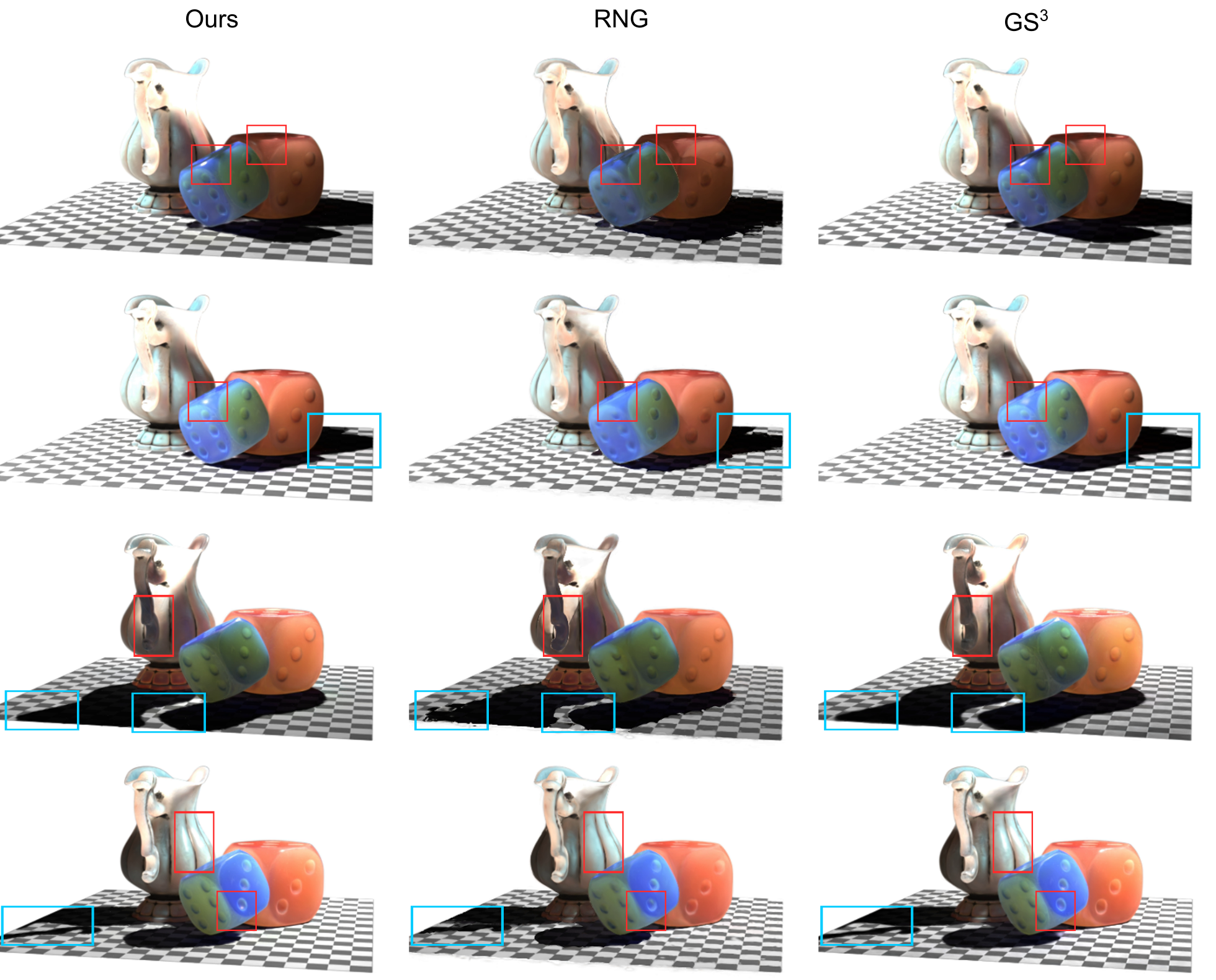}
    \caption{Qualitative relighting comparison on the \textit{Translucent} scene. 
Each row shows renderings from the same camera viewpoint under four novel lighting conditions. Our method is compared against RNG~\citep{fan_rng_2025} and GS$^3$~\citep{bi_gs3_2024}, demonstrating superior fidelity in reproducing light–material interactions.}
    \label{fig:comparison_relighting}
\end{figure}

Furthermore, we compare relighting results on the \textit{Translucent} scene with two OLAT-based Gaussian relighting methods, RNG~\citep{fan_rng_2025} and GS$^3$~\citep{bi_gs3_2024}, as shown in Fig.~\ref{fig:comparison_relighting}. The results demonstrate the superior relighting quality of our method, particularly in handling complex light–material interactions. For additional qualitative relighting results, please refer to the supplementary video.

In summary, our relighting analysis not only verifies the robustness of our approach under diverse lighting and viewing conditions, but also lays a fundamental basis for future research on controllable relighting, with broad applications in appearance editing, material-aware reconstruction, and immersive content creation.

\begin{table}[h]
\centering
\caption{Ablation study results on both real and synthetic datasets. The best/second-best results are colored in \colorbox[rgb]{1,0.8,0.8}{\strut red} / \colorbox[rgb]{1,0.9,0.8}{\strut orange}.}
\label{tab:ablation_study_additional-scene}
\small
\resizebox{\textwidth}{!}{
\begin{tabular}{l c | c c | c c c c | c c | c}
\toprule
\textbf{Dataset} & & \multicolumn{2}{c|}{\textbf{NRHints}} 
& \multicolumn{4}{c|}{\textbf{GS$^{3}$}} 
& \multicolumn{2}{c|}{\textbf{SSS-GS}}
& \textbf{Average} \\
\cline{3-10}
\textbf{Scenes} & & Pixiu & Fish & Translucent & FurBall & Lego & Hotdog & Bunny & Dragon & \\
\midrule
\multicolumn{11}{c}{\textbf{PSNR} $\uparrow$} \\
\midrule
\multirow{2}{*}{A: Diff} 
& Train & 20.2869 & 23.7003	& 16.6152 & 18.5678 & 20.6157 & 18.4717 & 21.3548 & 27.7851 & 20.9247 \\
& Test  & 20.1878 & 24.6146 & 15.3146 & 17.4010 & 17.6430 & 16.4279 & 21.3013 & 27.7071 & 20.0747 \\
\midrule
\multirow{2}{*}{B: D + S} 
& Train & 20.7321 & 24.8949 & 16.6980 & 18.3871 & 20.6891 & 18.5180 & 21.4677 & 27.9589 & 21.1682 \\
& Test  & 20.5692 & 25.3627 & 15.3200 & 17.1781 & 17.6290 & 16.4429 & 21.2236 & 27.6713 & 20.1746 \\
\midrule
\multirow{2}{*}{C: D + S + SSS} 
& Train & 25.1545 & 24.8573 & 26.4267 & 18.3845 & 25.6734 & 18.4628 & 22.7074 & 28.8883 & 23.8194 \\
& Test  & 24.8100 & 25.3275 & 25.6081 & 17.1655 & 22.8370 & 16.3730 & 22.3572 & 28.5486 & 22.8784 \\
\midrule
\multirow{2}{*}{D: Full (Ours)} 
& Train & \cellcolor[rgb]{1,0.8,0.8} 33.6065 & \cellcolor[rgb]{1,0.8,0.8} 32.0748 & \cellcolor[rgb]{1,0.8,0.8} 32.6058 & \cellcolor[rgb]{1,0.8,0.8} 35.4793 & \cellcolor[rgb]{1,0.8,0.8} 31.1434 & \cellcolor[rgb]{1,0.8,0.8} 32.4901 & \cellcolor[rgb]{1,0.8,0.8} 40.7672 & \cellcolor[rgb]{1,0.8,0.8} 39.3646 & \cellcolor[rgb]{1,0.8,0.8} 34.6915 \\
& Test  & \cellcolor[rgb]{1,0.8,0.8} 31.1213 & \cellcolor[rgb]{1,0.9,0.8} 31.1646 & \cellcolor[rgb]{1,0.8,0.8} 32.3919 & \cellcolor[rgb]{1,0.8,0.8} 35.1639 & \cellcolor[rgb]{1,0.8,0.8} 30.4664 & \cellcolor[rgb]{1,0.8,0.8} 32.1330 & \cellcolor[rgb]{1,0.8,0.8} 37.2270 & \cellcolor[rgb]{1,0.8,0.8} 36.6325 & \cellcolor[rgb]{1,0.8,0.8} 33.2876 \\
\midrule
\multirow{2}{*}{E: Full - S} 
& Train & \cellcolor[rgb]{1,0.9,0.8} 32.3487 & 28.0442 & \cellcolor[rgb]{1,0.9,0.8} 30.9386 & \cellcolor[rgb]{1,0.9,0.8} 33.7976 & \cellcolor[rgb]{1,0.9,0.8} 30.0839 & 30.0792 & 34.8697 & 36.3361 & 32.0623 \\
& Test  & \cellcolor[rgb]{1,0.9,0.8} 30.5952 & 28.2837 & \cellcolor[rgb]{1,0.9,0.8} 30.6761 & \cellcolor[rgb]{1,0.9,0.8} 34.0481 & \cellcolor[rgb]{1,0.9,0.8} 29.2294 & 30.4379 & 33.9633 & 35.8654 & 31.6374 \\
\midrule
\multirow{2}{*}{F: Full - SSS} 
& Train & 31.5332 & \cellcolor[rgb]{1,0.9,0.8} 31.9130 & 29.8005 & 32.6455 & 30.0656 & \cellcolor[rgb]{1,0.9,0.8} 31.5471 & \cellcolor[rgb]{1,0.9,0.8} 35.7335 & \cellcolor[rgb]{1,0.9,0.8} 37.5666 & \cellcolor[rgb]{1,0.9,0.8} 32.6006 \\
& Test  & 30.4095 & \cellcolor[rgb]{1,0.8,0.8} 31.1763 & 30.6253 & 33.8019 & 29.0048 & \cellcolor[rgb]{1,0.9,0.8} 32.0991 & \cellcolor[rgb]{1,0.9,0.8} 33.6715 & \cellcolor[rgb]{1,0.9,0.8} 35.9968 & \cellcolor[rgb]{1,0.9,0.8} 32.0982 \\
\bottomrule
\toprule
\multirow{2}{*}{H: Joint} 
& Train & 32.5452 & 31.4431 & 29.3508 & 32.5538 & 29.8266 & 30.5236 & 34.0744 & 37.0531 & 32.1713 \\
& Test  & 31.0880 & 30.4943 & 30.6756 & 33.6832 & 29.1154 & 31.7083 & 31.7589 & 35.2621 & 31.7232 \\
\midrule
\multirow{2}{*}{I: Prog. Phys (Ours)} 
& Train & \cellcolor[rgb]{1,0.8,0.8} 33.6065 & \cellcolor[rgb]{1,0.8,0.8} 32.0748 & \cellcolor[rgb]{1,0.8,0.8} 32.6058 & \cellcolor[rgb]{1,0.8,0.8} 35.4793 & \cellcolor[rgb]{1,0.8,0.8} 31.1434 & \cellcolor[rgb]{1,0.8,0.8} 32.4901 & \cellcolor[rgb]{1,0.8,0.8} 40.7672 & \cellcolor[rgb]{1,0.8,0.8} 39.3646 & \cellcolor[rgb]{1,0.8,0.8} 34.6915 \\
& Test  & \cellcolor[rgb]{1,0.8,0.8} 31.1213 & \cellcolor[rgb]{1,0.9,0.8} 31.1646 & \cellcolor[rgb]{1,0.8,0.8} 32.3919 & \cellcolor[rgb]{1,0.8,0.8} 35.1639 & \cellcolor[rgb]{1,0.8,0.8} 30.4664 & \cellcolor[rgb]{1,0.8,0.8} 32.1330 & \cellcolor[rgb]{1,0.8,0.8} 37.2270 & \cellcolor[rgb]{1,0.8,0.8} 36.6325 & \cellcolor[rgb]{1,0.8,0.8} 33.2876 \\
\midrule
\multirow{2}{*}{J: Prog. NonPhys} 
& Train & 32.5606 & 31.3321 & 29.6999 & \cellcolor[rgb]{1,0.9,0.8} 32.8201 & \cellcolor[rgb]{1,0.9,0.8} 30.7004 & 30.9712 & 35.9882 & \cellcolor[rgb]{1,0.9,0.8} 37.6112 & 32.7105 \\
& Test  & \cellcolor[rgb]{1,0.9,0.8} 31.0973 & 30.2406 & 30.5955 & 33.9654 & \cellcolor[rgb]{1,0.9,0.8} 29.4718 & 31.2575 & 34.0090 & \cellcolor[rgb]{1,0.9,0.8} 36.0222 & 32.0824 \\
\midrule
\multirow{2}{*}{K: Prog. Merge} 
& Train & \cellcolor[rgb]{1,0.9,0.8} 33.3438 & \cellcolor[rgb]{1,0.9,0.8} 32.0247 & \cellcolor[rgb]{1,0.9,0.8} 31.7351 & 32.6704 & 30.2933 & \cellcolor[rgb]{1,0.9,0.8} 31.4903 & \cellcolor[rgb]{1,0.9,0.8} 39.1104 & 37.5251 & \cellcolor[rgb]{1,0.9,0.8} 33.5241 \\
& Test  & 31.0486 & \cellcolor[rgb]{1,0.8,0.8} 31.2827 & \cellcolor[rgb]{1,0.9,0.8} 31.8883 & \cellcolor[rgb]{1,0.9,0.8} 33.9799 & 29.2607 & \cellcolor[rgb]{1,0.9,0.8} 32.0399 & \cellcolor[rgb]{1,0.9,0.8} 36.6183 & 35.8016 & \cellcolor[rgb]{1,0.9,0.8} 32.7400 \\
\bottomrule
\end{tabular}
}
\end{table}

\begin{figure*}[h]
    \centering
    \includegraphics[width=1\linewidth]    {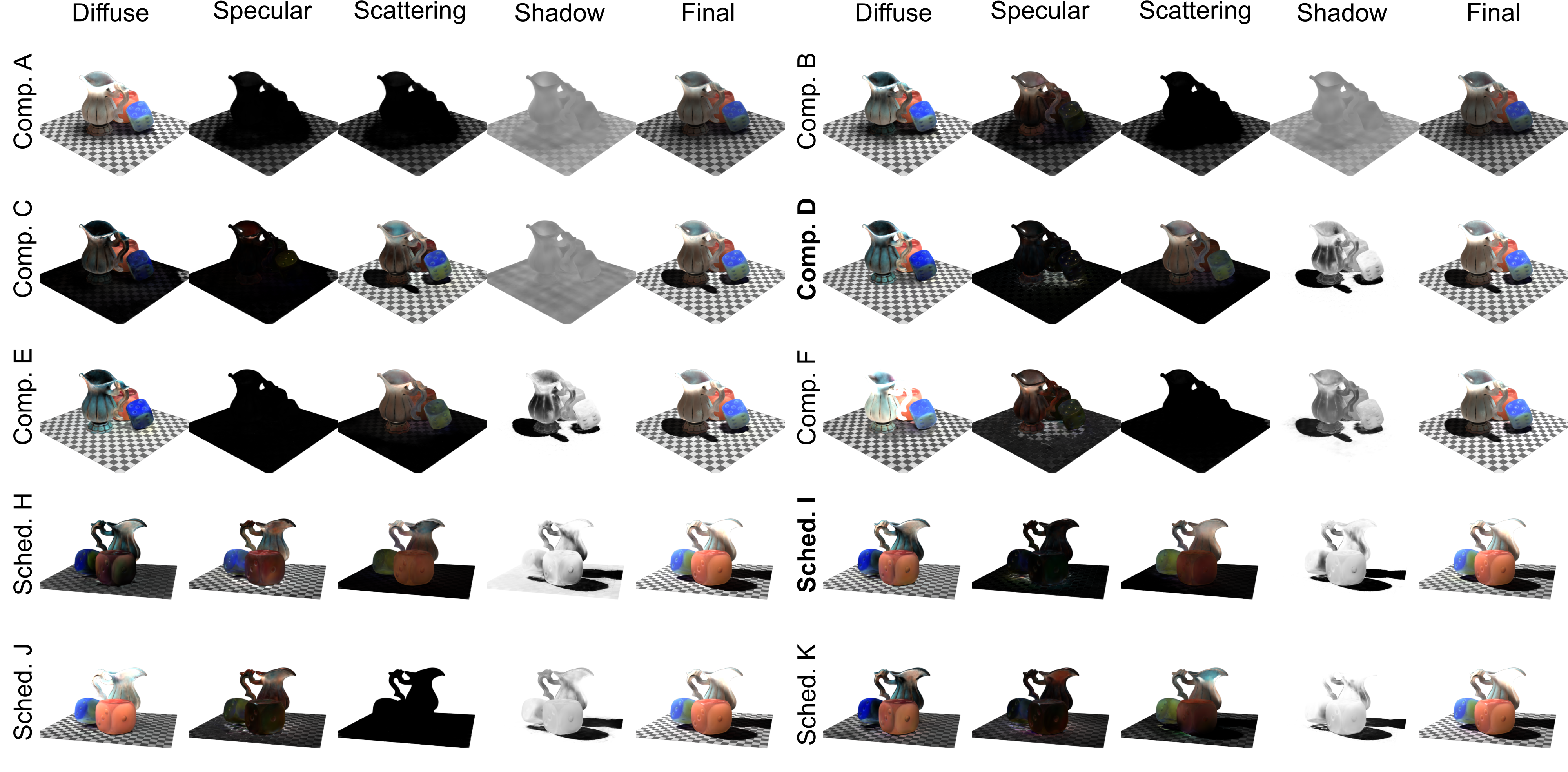}
    \vspace{-0.2in}
    \caption{Visualization of reconstructed components under different reflectance decompositions and training schedules on the \textit{Translucent} scene from the 
GS$^3$ synthetic dataset. Top: six reflectance compositions (Comp. A-F); Bottom: four training schedules (Sched. H-K).}
    \label{fig:ablationstudy_translucent}
\end{figure*}

\section{\revise{Additional Ablation Study}}
\refstepcounter{subsection}
\paragraph{Reflectance Components and Training Schedule.}
\label{appendix:additional-ablation-study-components-schedules}
In addition to the Pixiu example presented in the main paper (Fig. \ref{fig:ablationstudy_pixiu} and Tab. \ref{tab:ablation_study}), we provide expanded ablation results across more real and synthetic scenes to illustrate the generality of our observations. For the reflectance components, we report per-scene comparisons for diffuse, specular, subsurface scattering (SSS), and shadow terms, highlighting consistent trends in directional visibility, highlight formation, and translucent appearance. We also include additional per-scene evaluations, following our progressive optimization strategy, for the training schedule. The supplementary Tab. \ref{tab:ablation_study_additional-scene} and visualizations Fig.\ref{fig:ablationstudy_translucent} confirm that the behaviors observed in the main paper hold robustly across diverse materials and lighting conditions.

\begin{table}[h]
\centering
\caption{Quantitative comparison when applying the shadow term to the SSS component.}
\label{tab:ablationstudy_shadow-on-sss}
\small
\resizebox{\textwidth}{!}{
\begin{tabular}{l c | c c c | c c c c | c c | c}
\toprule
\textbf{Dataset} & & \multicolumn{3}{c|}{\textbf{NRHints}} 
& \multicolumn{4}{c|}{\textbf{GS$^{3}$}} 
& \multicolumn{2}{c|}{\textbf{SSS-GS}}
& \textbf{Average} \\
\cline{3-11}
\textbf{Scenes} & & Pixiu & Fish & FurScene & Translucent & FurBall & Lego & Hotdog & Bunny & Dragon & \\
\midrule
\multicolumn{12}{c}{\textbf{PSNR} $\uparrow$} \\
\midrule
\multirow{2}{*}{Shadow-on-SSS} 
& Train & 32.1850 & 31.7531 & 31.6750 & 30.8066 & 35.4219 & 31.1002 & 32.1094 & 35.9068 & 37.3346 & 33.1436 \\
& Test  & 30.9756 & 31.1206 & 30.6575 & 30.0740 & 34.9420 & 30.3487 & 31.5272 & 33.6508 & 35.8476 & 32.1271 \\
\midrule
\multirow{2}{*}{Ours} 
& Train & 33.6065 & 32.0748 & 31.7846 & 32.6058 & 35.4793 & 31.1434 & 32.4901 & 40.7672 & 39.3646 & \cellcolor[rgb]{1,0.8,0.8} 34.3685 \\
& Test  & 31.1213 & 31.1646 & 30.7349 & 32.3919 & 35.1639 & 30.4664 & 32.1330 & 37.2270 & 36.6325 & \cellcolor[rgb]{1,0.8,0.8} 33.0039 \\
\midrule
\multicolumn{12}{c}{\textbf{SSIM} $\uparrow$} \\
\midrule
\multirow{2}{*}{Shadow-on-SSS} 
& Train & 0.9484 & 0.9337 & 0.9570 & 0.9761 & 0.9768 & 0.9701 & 0.9755 & 0.9841 & 0.9815 & 0.9670 \\
& Test  & 0.9432 & 0.9252 & 0.9510 & 0.9751 & 0.9732 & 0.9562 & 0.9721 & 0.9762 & 0.9760 & 0.9609 \\
\midrule
\multirow{2}{*}{Ours} 
& Train & 0.9524 & 0.9363 & 0.9576 & 0.9835 & 0.9776 & 0.9706 & 0.9776 & 0.9922 & 0.9874 & \cellcolor[rgb]{1,0.8,0.8} 0.9705 \\
& Test  & 0.9452 & 0.9260 & 0.9518 & 0.9823 & 0.9733 & 0.9570 & 0.9743 & 0.9859 & 0.9789 & \cellcolor[rgb]{1,0.8,0.8} 0.9638 \\
\midrule
\multicolumn{12}{c}{\textbf{LPIPS} $\downarrow$} \\
\midrule
\multirow{2}{*}{Shadow-on-SSS} 
& Train & 0.0797 & 0.0811 & 0.0695 & 0.0278 & 0.0502 & 0.0336 & 0.0305 & 0.0285 & 0.0236 & 0.0472 \\
& Test  & 0.0825 & 0.0880 & 0.0726 & 0.0277 & 0.0450 & 0.0424 & 0.0327 & 0.0360 & 0.0285 & 0.0506 \\
\midrule
\multirow{2}{*}{Ours} 
& Train & 0.0751 & 0.0779 & 0.0690 & 0.0201 & 0.0482 & 0.0331 & 0.0293 & 0.0113 & 0.0157 & \cellcolor[rgb]{1,0.8,0.8} 0.0422 \\
& Test  & 0.0791 & 0.0855 & 0.0724 & 0.0200 & 0.0442 & 0.0419 & 0.0326 & 0.0179 & 0.0240 & \cellcolor[rgb]{1,0.8,0.8} 0.0464 \\
\bottomrule
\end{tabular}
}
\end{table}

\begin{figure}[h]
\centering
\includegraphics[width=\linewidth]{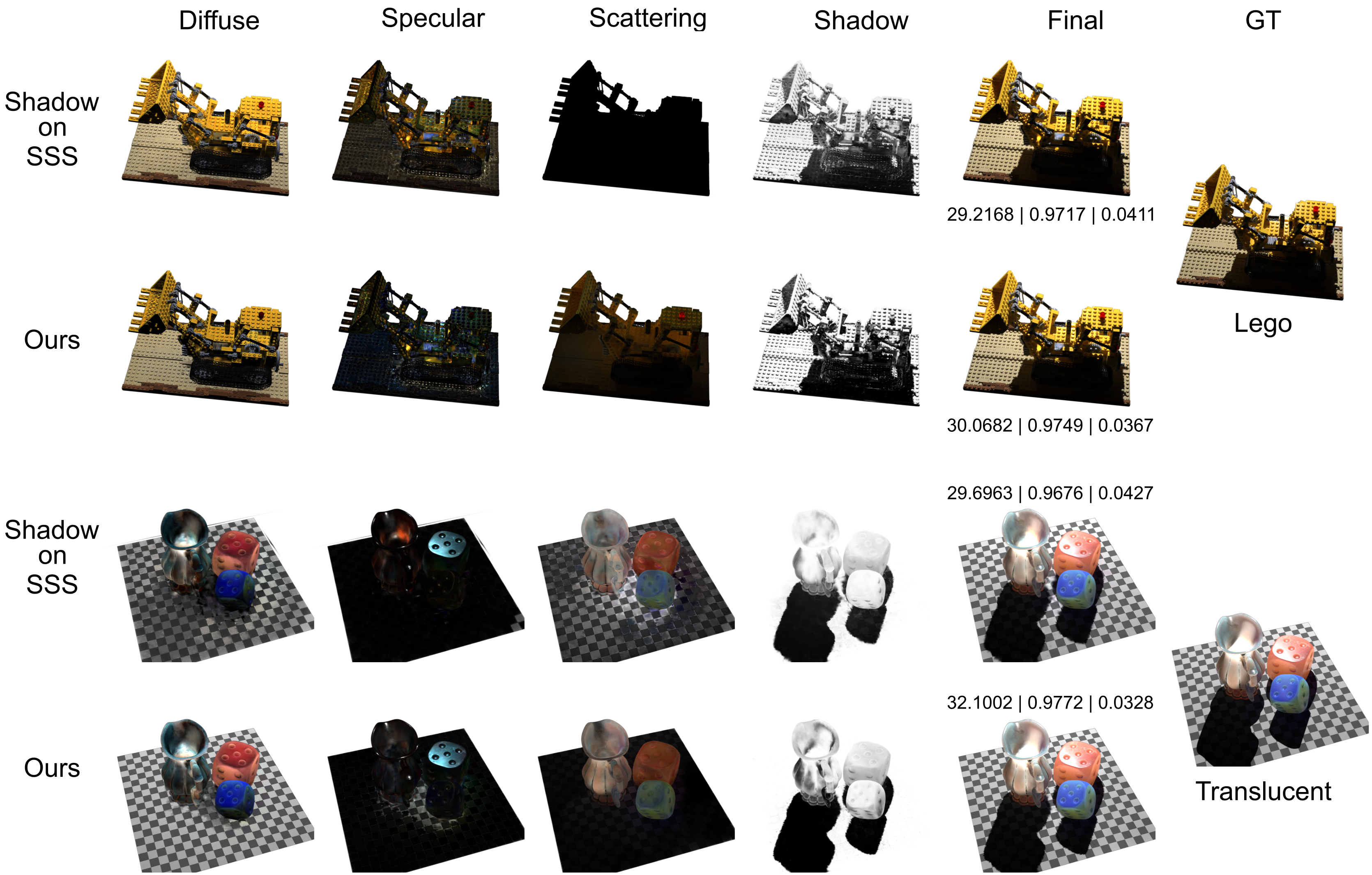}
\caption{
Qualitative comparison for the Shadow–SSS interaction ablation study. Applying the shadow term to the SSS component leads to over-darkened scattering, loss of back-lit translucency, and reduced soft shading. Our formulation preserves translucent appearance while maintaining consistent shadow behavior.
}
\label{fig:ablationstudy_shadow-on-sss}
\end{figure}

\refstepcounter{subsection}
\paragraph{Shadow–SSS Interaction.}
\label{appendix:shadow-sss-interaction}
To examine how the shadow term $S(\mathbf{x})$ interacts with subsurface scattering (SSS), we evaluate an alternative shading variant in which the shadow term is also applied to the SSS component. For clarity, we restate the shading model used in the main paper (Eq.~\ref{eq:shading_equation}):
\begin{equation}
\text{Ours: } 
(c_d f_d + c_s f_s) \cdot S(\mathbf{x}) + c_{sss} f_{sss}
\end{equation}
where the shadow term modulates only the diffuse and specular contributions. In the ablation, we instead apply the shadow term to all components, resulting in
\begin{equation}
\text{Shadow-on-SSS: }
(c_d f_d + c_s f_s + c_{sss} f_{sss}) \cdot S(\mathbf{x})
\end{equation}

This design highlights the distinction between surface occlusion and volumetric diffusion. The SSS module is directly supervised by multi-view images, which already account for visibility and attenuation effects. Introducing $S(\mathbf{x})$ to the SSS component results in a secondary attenuation that reduces back-lit translucency and causes overly dark scattering. Our ablation study confirms that this approach leads to lower reconstruction quality across translucent, semi-translucent, and opaque scenes.

Tab.~\ref{tab:ablationstudy_shadow-on-sss} reports per-scene metrics. The “Shadow-on-SSS” variant performs worse in all categories, with the largest differences observed in scenes that contain strong subsurface transport. Fig.~\ref{fig:ablationstudy_shadow-on-sss} shows representative qualitative results that demonstrate the loss of translucency and soft scattering when the shadow term is applied to the SSS component.

\section{\revise{Additional Relighting Baselines}}
\label{appendix:additional-relighting-baselines}
This section provides additional comparisons on the Synthetic OLAT dataset and further clarifies the modeling differences among recent relighting approaches. Existing relighting methods based on NeRF or 3D Gaussian Splatting generally differ in how illumination is parameterized and reconstructed, which directly determines their behavior under point-light relighting.

\paragraph{Relighting under unknown illumination.}
The first class of datasets contains scenes captured under \emph{unknown} and often complex illumination, such as outdoor environments or indoor scenes dominated by global illumination. Methods in this category, including TensoIR~\citep{jin_tensoir_2023}, R3DG~\citep{gao_relightable_2024}, IRGS~\citep{gu_irgs_2025}, and GI-GS~\citep{chen_gi-gs_2024}, must jointly infer surface reflectance, geometry, and an environment map from the observed radiance. Because the incoming light distribution is not provided, these approaches rely on explicit or residual global-illumination modeling (e.g., multi-bounce shading, occlusion volumes, or deferred visibility terms) to explain indirect energy that cannot be deduced from direct lighting alone. The recovered illumination is typically represented as a low-frequency environment map, making these methods effective for ambient relighting but fundamentally limited in reproducing the high-frequency, spatially localized behavior characteristic of point-light transport.

\paragraph{Relighting under known illumination.}
The second class of datasets follows a controlled One-Light-at-a-Time (OLAT) protocol, where each training view is illuminated by a \emph{known} single point light with calibrated position. Approaches such as GS$^{3}$~\citep{bi_gs3_2024}, RNG~\citep{fan_rng_2025}, and ours leverage this setting, which provides explicit per-light supervision and cleanly separates geometry, BRDF, and illumination. Unlike unknown-light datasets, OLAT observations directly reveal high-frequency shading cues, including directional visibility, sharp-to-soft shadow transitions, and localized subsurface transport, so explicit global-illumination terms become unnecessary. In this setting, introducing residual multi-bounce components often leads to ambiguity by entangling reflectance and transport. Instead, OLAT-oriented methods focus on accurately modeling direct point-light transport; our approach follows this paradigm using a physically structured Gaussian-domain shading model that captures both fine-scale direct effects and the small, naturally occurring low-frequency residual energy in controlled environments.

\paragraph{Expanded baselines.}
To broaden the relighting comparison, we additionally evaluate TensoIR and R3DG on the GS$^{3}$ Synthetic OLAT dataset. Both methods are re-trained using their official implementations. Since these approaches rely on environment-map estimation, they must explain point-light observations using low-frequency illumination representations. This mismatch introduces ambiguity and typically manifests as blurred shadows, reduced directional contrast, and attenuated high-frequency shading. Quantitative results in Tab.~\ref{tab:quantitative_synthetic} and Tab.~\ref{tab:quantitative_synthetic_relighting} show that OLAT-targeted methods consistently outperform unknown-lighting relighting models, and our method achieves the highest accuracy across all scenes and metrics.

\begin{table*}[h]
    \centering
    \caption{Quantitative comparison results with expanded relighting baselines: TensoIR and R3DG. The best results are colored in \colorbox[rgb]{1,0.8,0.8}{\strut red}.} 
    \vspace{-0.1in}
    \label{tab:quantitative_synthetic_relighting}

    \resizebox{\linewidth}{!}{
    \begin{tabular}{lc|c|c|c|c|c|c|c}
    \hline
    \multicolumn{2}{c|}{\diagbox{Method}{Dataset}} & Translucent & AnisoMetal & Drums & FurBall & Hotdog & Lego & Average \\
    \hline

    \multicolumn{9}{c}{\textbf{PSNR} $\uparrow$} \\
    \midrule
    \multirow{2}{*}{TensoIR} 
        & Train & 16.9800 & 18.1800 & 26.4400 & 20.1600 & 17.1900 & 17.5200 & 19.4117 \\ 
        & Test & 15.9997 & 16.8545 & 24.8974 & 20.0509 & 17.1500 & 17.0317 & 18.6640 \\ 
    \midrule

    \multirow{2}{*}{R3DG} 
        & Train & 17.1815 & 18.1650 & 27.3024 & 21.6486 & 19.5616 & 19.4608 & 20.5533 \\ 
        & Test & 16.4664 & 17.1060 & 25.0378 & 20.1443 & 17.1896 & 16.2879 & 18.7053 \\ 
    \midrule

    \multirow{2}{*}{Ours} 
        & Train & \cellcolor[rgb]{1,0.8,0.8}32.6058 
        & \cellcolor[rgb]{1,0.8,0.8}31.1077 
        & \cellcolor[rgb]{1,0.8,0.8}34.2448 
        & \cellcolor[rgb]{1,0.8,0.8}35.4793 
        & \cellcolor[rgb]{1,0.8,0.8}32.4901 
        & \cellcolor[rgb]{1,0.8,0.8}31.1434 
        & \cellcolor[rgb]{1,0.8,0.8}32.8452 \\

        & Test & \cellcolor[rgb]{1,0.8,0.8}32.3919 
        & \cellcolor[rgb]{1,0.8,0.8}30.0448 
        & \cellcolor[rgb]{1,0.8,0.8}33.5514 
        & \cellcolor[rgb]{1,0.8,0.8}35.1639 
        & \cellcolor[rgb]{1,0.8,0.8}32.1330 
        & \cellcolor[rgb]{1,0.8,0.8}30.4664 
        & \cellcolor[rgb]{1,0.8,0.8}32.2919 \\
    \hline

    \end{tabular}
    }
\end{table*}

These expanded comparisons demonstrate that relighting methods designed for unknown or environment-map illumination do not perform well in the point-light OLAT scenario, where high-frequency directional cues are essential. In contrast, known-light approaches—particularly those that incorporate physically structured transport modeling—achieve significantly more accurate and consistent results. Our method attains state-of-the-art performance on the GS$^{3}$ Synthetic OLAT dataset.

\section{\revise{Analysis of Shadow}}
\refstepcounter{subsection}
\paragraph{Shadow Pipeline.}
This section analyzes the shadow term and visualizes the behavior of our visibility formulation. Although a point light theoretically produces a sharp umbra boundary, our continuous transmittance model yields smooth and geometry-aware transitions. The per-ray transmittance $v_i$ accumulates attenuation along each shadow ray, and its spatial variation across neighboring rays naturally induces soft penumbra regions. Aggregating these ray-wise values into a per-Gaussian coarse visibility $\hat v_g$ further smooths local discontinuities and captures how each Gaussian contributes to shadowing. The refined shadow $S(x)$ then maps these visibility cues to pixel-space shadow intensities, suppressing residual artifacts and producing stable, physically interpretable soft shadows. As shown in Fig.~\ref{fig:shadow_analysis_pipeline}, the progression from $v_i$ to $\hat v_g$ and finally to $S(x)$ illustrates how continuous visibility modeling produces coherent, geometry-consistent soft shadows under point-light illumination.

\begin{figure}[t]
    \centering
    \includegraphics[width=\linewidth]{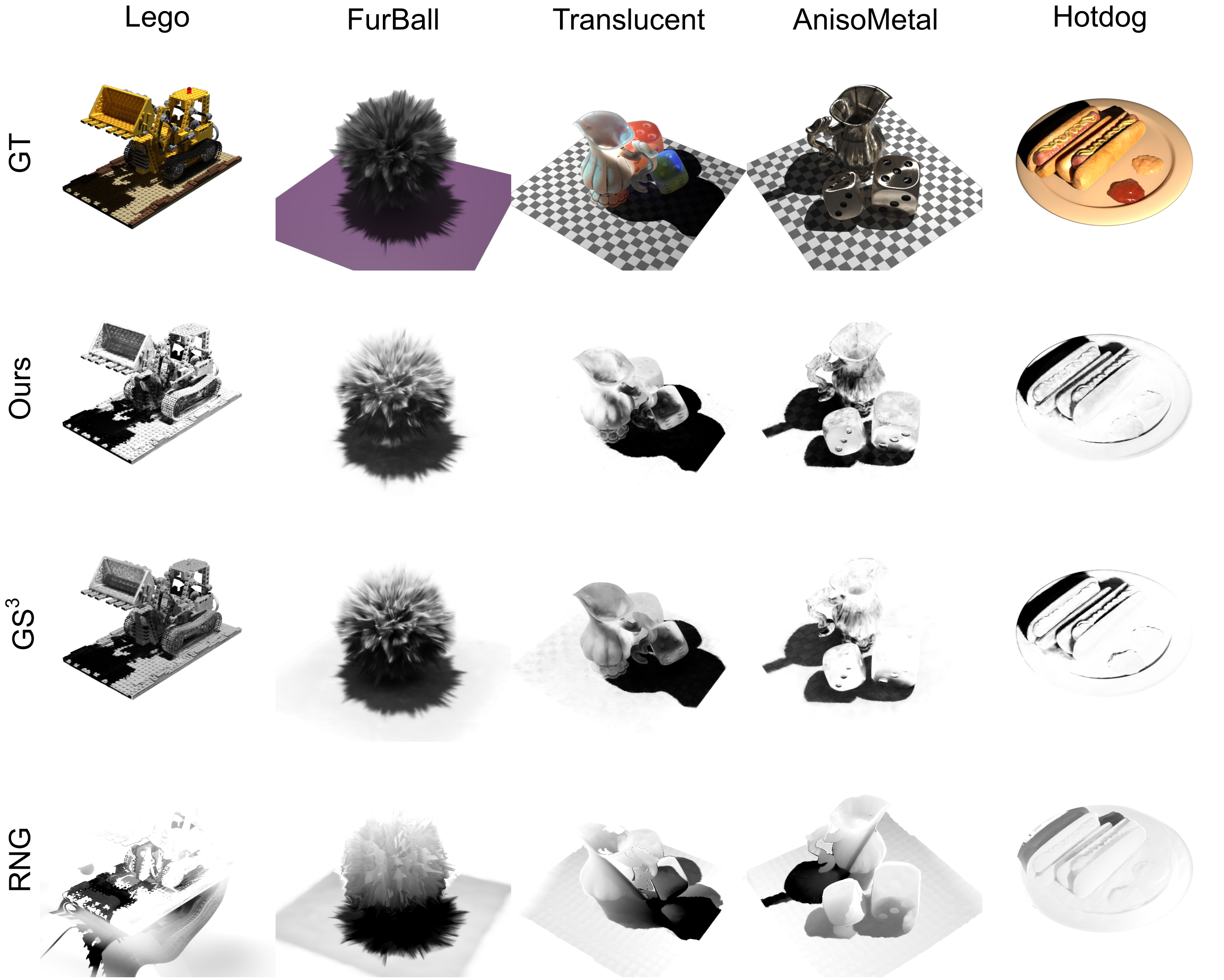}
    \caption{Comparison with screen-space shadow baselines. Top row: reference renderings. Second row: shadows produced by our method. Third and fourth rows: shadows from GS$^3$~\citep{bi_gs3_2024} and RNG~\citep{fan_rng_2025}, which both rely on screen-space opacity accumulation. }
    \label{fig:shadow_comparison_with_baseline}
\end{figure}

\refstepcounter{subsection}
\paragraph{Comparison with Screen-Space Shadow Baselines.}
\label{appendix:comparison-screen-space-shadow-baselines}
We additionally compare our formulation with screen-space opacity-accumulation strategies used in methods such as GS$^3$~\citep{bi_gs3_2024} and RNG~\citep{fan_rng_2025}, as illustrated in Fig.~\ref{fig:shadow_comparison_with_baseline}. Since these approaches compute shadowing after projection, they are highly sensitive to depth ordering and often exhibit unstable or overly sharp shadow boundaries, particularly around thin structures or regions with multi-layer occlusion. Screen-space accumulation also struggles to maintain consistency under viewpoint changes, as small perturbations in projected splat order can produce flickering or discontinuities.

In contrast, our volumetric visibility formulation integrates attenuation along the light ray in 3D, independent of screen-space ordering. This yields smoother and more geometry-consistent transitions, stable penumbra behavior, and improved handling of dense or overlapping Gaussians and concave geometry. These comparisons highlight the advantages of continuous transmittance and emphasize the importance of modeling visibility at both the ray and Gaussian levels rather than relying solely on post-projection image-space accumulation.

\clearpage

\section{LLM Usage}

Throughout this study, we used LLMs only to assist with writing—correcting grammar and refining phrasing to improve clarity.

We \textbf{did not} use LLMs to search for or identify related works; all literature was found by the authors.

LLMs \textbf{did not }contribute to the intellectual development of the research.

\end{document}